\documentclass[journal]{new-aiaa}
\usepackage[utf8]{inputenc}
\usepackage{textcomp}
\usepackage{graphicx}
\graphicspath{{graphs/}}
\usepackage{amsmath}
\usepackage{siunitx}
\usepackage{longtable,tabularx}
\usepackage{booktabs}
\usepackage{array}
\usepackage{xcolor}
\usepackage{subcaption}

\definecolor{adsbblue}{RGB}{41,128,185}
\definecolor{audiogreen}{RGB}{39,174,96}
\definecolor{linkpurple}{RGB}{142,68,173}
\definecolor{jointorange}{RGB}{211,84,0}
\definecolor{darkgray}{RGB}{80,80,80}
\definecolor{voiceC}{HTML}{B5482E}
\definecolor{trajC}{HTML}{1F5C99}
\definecolor{jointC}{HTML}{5B4B8A}

\usepackage{placeins}

\hypersetup{
  colorlinks=true,
  linkcolor=black,
  citecolor=black,
  urlcolor=blue!60!black,
}

\usepackage{tikz}
\usetikzlibrary{shapes.geometric, arrows.meta, positioning, fit, calc, backgrounds}
\usetikzlibrary{arrows.meta, backgrounds, fit, calc, plotmarks}

\newcommand{\R}{\mathbb{R}}
\newcommand{\Xtraj}{\mathbf{X}}
\newcommand{\wav}{w}
\newcommand{\vemb}{\mathbf{v}}
\newcommand{\tremb}{\mathbf{r}}
\newcommand{\jemb}{\mathbf{j}}
\newcommand{\vproj}{\hat{\mathbf{v}}}
\newcommand{\tproj}{\hat{\mathbf{r}}}
\newcommand{\vnorm}{\tilde{\mathbf{v}}}
\newcommand{\tnorm}{\tilde{\mathbf{r}}}
\newcommand{\Ev}{\mathcal{E}_v}
\newcommand{\Et}{\mathcal{E}_t}
\newcommand{\Pv}{P_v}
\newcommand{\Pt}{P_t}
\newcommand{\fv}{f_v}
\newcommand{\ft}{f_t}
\newcommand{\Lloss}{\mathcal{L}}
\newcommand{\Mset}{\mathcal{M}}

\title{V2TATC: Joint Voice-Trajectory Embedding and Dataset for Air Traffic Controller Situational Awareness}

\author{Louis Brusset\footnote{Graduate Student, Mines Paris, PSL University; e-mail: \href{mailto:louis.brusset@mines-paris.org}{louis.brusset@mines-paris.org} (corresponding author).}}
\affil{Mines Paris--PSL University, Paris, 75006, France}

\author{Mathurin Petit\footnote{Graduate Student, \'Ecole Polytechnique, Institut Polytechnique de Paris.}}
\affil{\'{E}cole Polytechnique, Palaiseau, 91128, France}

\author{Jordan Kam\footnote{Ph.D. Student, Department of Aerospace Engineering, California Institute of Technology, AIAA Student Member.}}
\affil{California Institute of Technology, Pasadena, CA, 91125, USA}

\author{Alexandre M. Bayen\footnote{Liao-Cho Endowed Chair Professor, Department of Electrical Engineering and Computer Sciences, University of California, Berkeley.}}
\affil{University of California, Berkeley, Berkeley, CA, 94720, USA}

\begin{document}

\maketitle

\begin{abstract}
As air traffic volumes in the National Airspace System continue to expand, in particular at low altitude, the need for scalable decision support tools used by air traffic controllers will also require more development. This article introduces Voice-to-Trajectory for Air Traffic Control, a joint voice communication-flight trajectory data embedding framework, that can be a component of situational awareness in congested airspaces, and assist the development of tools for ATC as they reason in real-time over Automatic Dependent Surveillance-Broadcast trajectories, or the intent expressed by pilots in natural language. We show that these data modalities are not independent and represent a common physical referent: an aircraft flying through the airspace. V2TATC maps a voice instruction and the trajectory of the addressed aircraft to nearby points in a single latent space that can be queried in both directions. It combines a self-supervised trajectory encoder, a frozen speech encoder, a contrastive joint embedding, and a bijective lifting via normalizing flows. We demonstrate V2TATC's effectiveness on the San Francisco Bay Area, for its concentration of major airports, and its mix of commercial and general aviation traffic. Lastly, we release a novel paired voice-trajectory dataset, and report experiments on cross-modal retrieval, ablations, and latent-space analysis.
\end{abstract}

\newpage

\section*{Nomenclature}
\noindent(All quantities are dimensionless unless units are given.)
{\renewcommand\arraystretch{1.0}
\noindent\begin{longtable*}{@{}l @{\quad=\quad} l@{}}
$\Xtraj$         & trajectory window, $T \times F$ array of surveillance samples \\
$x_t$            & feature vector at timestep $t$, with $x_t \in \R^F$ \\
$\wav$           & raw voice waveform sampled at \SI{16}{\kilo\hertz} \\
$\phi, \lambda, \theta$  & geodetic latitude and longitude, true track angle, rad \\
$v, \dot{h}$     & ground speed and vertical rate, m\,s$^{-1}$ \\
$d_v, d_t, d_j$  & voice, trajectory, and joint embedding dimensions ($d_v = 1280$, $d_t = 1792$, $d_j = 1024$)\\
$\Ev, \Et$       & voice and trajectory encoders \\
$\Pv, \Pt$       & voice and trajectory projectors \\
$\fv, \ft$       & voice and trajectory normalizing flows in $\R^{d_j}$ \\
$\vemb, \tremb$  & voice and trajectory embeddings, $\vemb \in \R^{d_v}$, $\tremb \in \R^{d_t}$ \\
$\vproj, \tproj$ & projected voice and trajectory embeddings, $\vproj, \tproj \in \R^{d_j}$ \\
$\vnorm, \tnorm$ & $\ell_2$-normalized projected voice and trajectory embeddings \\
$\jemb$          & joint-space representation, $\jemb \in \R^{d_j}$ \\
$\Mset$          & set of masked timestep indices \\
$\Lloss_{\bullet}$ & loss term (subscript denotes its role) \\
$\lambda_{\bullet}$ & weight term in a loss (subscript denotes its role) \\
$S_{ij}$         & cosine similarity between voice $i$ and trajectory $j$ \\
$\tau$           & contrastive temperature \\
$K$              & retrieval cutoff used in Recall@$K$ \\
\end{longtable*}}
\addtocounter{table}{-1}
\section{Introduction}

\subsection*{Operational needs in the National Airspace System}
\lettrine{A}{s} traffic congestion density in the \textit{National Airspace System} (NAS) continues to increase, \textit{decision support tools} (DST) used by air traffic controllers must also become increasingly capable of supporting this growth in high-stakes operations \cite{chi2023review}. \textit{Air Traffic Control} (ATC) requires real-time reasoning over multiple streams of multimodal information in sequenced safety-critical scenarios. Without always realizing it, a controller managing a congested terminal airspace continuously analyzes 3D trajectories from \emph{Automatic Dependent Surveillance--Broadcast (ADS-B)}, the intent expressed by pilots in natural language over radio, environmental conditions reported by \emph{Meteorological Aerodrome Reports (METAR)}, digital infrastructure of the \textit{Federal Aviation Administration (FAA)} (\textit{navigation aids (navaids)}, \textit{VHF Omnidirectional Range (VORs)}, etc.), and many other auxiliary signals \cite{darrell2026automated}. This ATC data represents a common physical entity, an aircraft flying through the airspace, and it is precisely the implicit relationship between them that allows a human controller to maintain situational awareness and ensure safe separation. Most existing DSTs, by contrast, process these streams in isolation. For example, speech recognition pipelines transcribe radio audio into text and further modules recover speaker roles and turn boundaries from those transcripts \cite{helmke2024asru,zuluaga2022bertraffic}. Trajectory prediction systems operate on ADS-B features alone, whether sourced from operational surveillance or from large-scale networks such as OpenSky \cite{schafer2014opensky}, and forecast future positions with recurrent or attention-based models \cite{zhao2019lstm,guo2023flightbert}. Weather is used through a separate display, leaving its integration to the controller \cite{chi2023review}. The semantic correspondence between a controller's verbal instruction and the resulting maneuver, which is the very basis of safe and efficient operations, is left implicit, and processed by the human.

This article summarizes results that demonstrate that this implicit relationship can be learned. We present \textit{Voice-to-Trajectory for Air Traffic Control} (V2TATC), a framework in which voice transmissions and ADS-B trajectories live in a single representation space, so that a voice transmission and the trajectory of the addressed aircraft are mapped to nearby points, and so that the mapping can be queried in both directions. The idea is inspired by recent progress in multi-modal joint embeddings, in particular the contrastive language--image alignment introduced by \textit{Contrastive Language-Image Pre-training} (CLIP)~\cite{radford2021clip} and its successors in vision--language~\cite{jia2021align,li2022blip}, and by the maturity of large-scale speech encoders such as Whisper~\cite{radford2023whisper} and Wav2Vec~2.0~\cite{baevski2020wav2vec2}. These models share a common component, namely a pre-trained per-modality encoder followed by a lightweight alignment head trained on paired data using a contrastive objective. We adopt the same approach with two specificities that matter for ATC. First, the alignment target is not an open-ended natural-language description. It is a structured time series of several features broadcast at a roughly constant rate. This calls for a specialized trajectory encoder that captures motion dynamics, which we obtain through a \textit{Masked Autoencoder} (MAE) objective~\cite{he2022mae}. Second, retrieval alone is not enough, because a DST may eventually need to generate a candidate trajectory from a voice instruction, or vice versa. This calls for an invertible alignment, which we obtain by training two normalizing flows~\cite{dinh2017realnvp,kingma2018glow} on top of the contrastive joint space.

The joint embedding is intended as a building block for DSTs rather than as an end-user application by itself. A small set of operations on top of the joint space already maps to concrete controller needs. Selecting an aircraft on the surveillance display and recovering the corresponding recent voice transmissions (\emph{trajectory~$\to$~voice} retrieval) supports rapid context recovery during handovers and high-workload situations. The symmetric operation, going from a clipped voice transmission to a candidate trajectory (\emph{voice~$\to$~trajectory}), supports listening tools that help a supervisor verify which aircraft is being addressed. Inconsistency between a voice instruction and the corresponding trajectory in the joint space provides a straightforward signal for anomaly detection, including missed acknowledgments and non-compliant maneuvers. Finally, the bijective lifting opens the door to generative decision support: short-horizon trajectory forecasts conditioned on voice instructions, plausible-instruction synthesis for simulator training, and conflict probes that explore counterfactual maneuvers.

\begin{figure}[tp]
\centering
\includegraphics[width=0.9\textwidth]{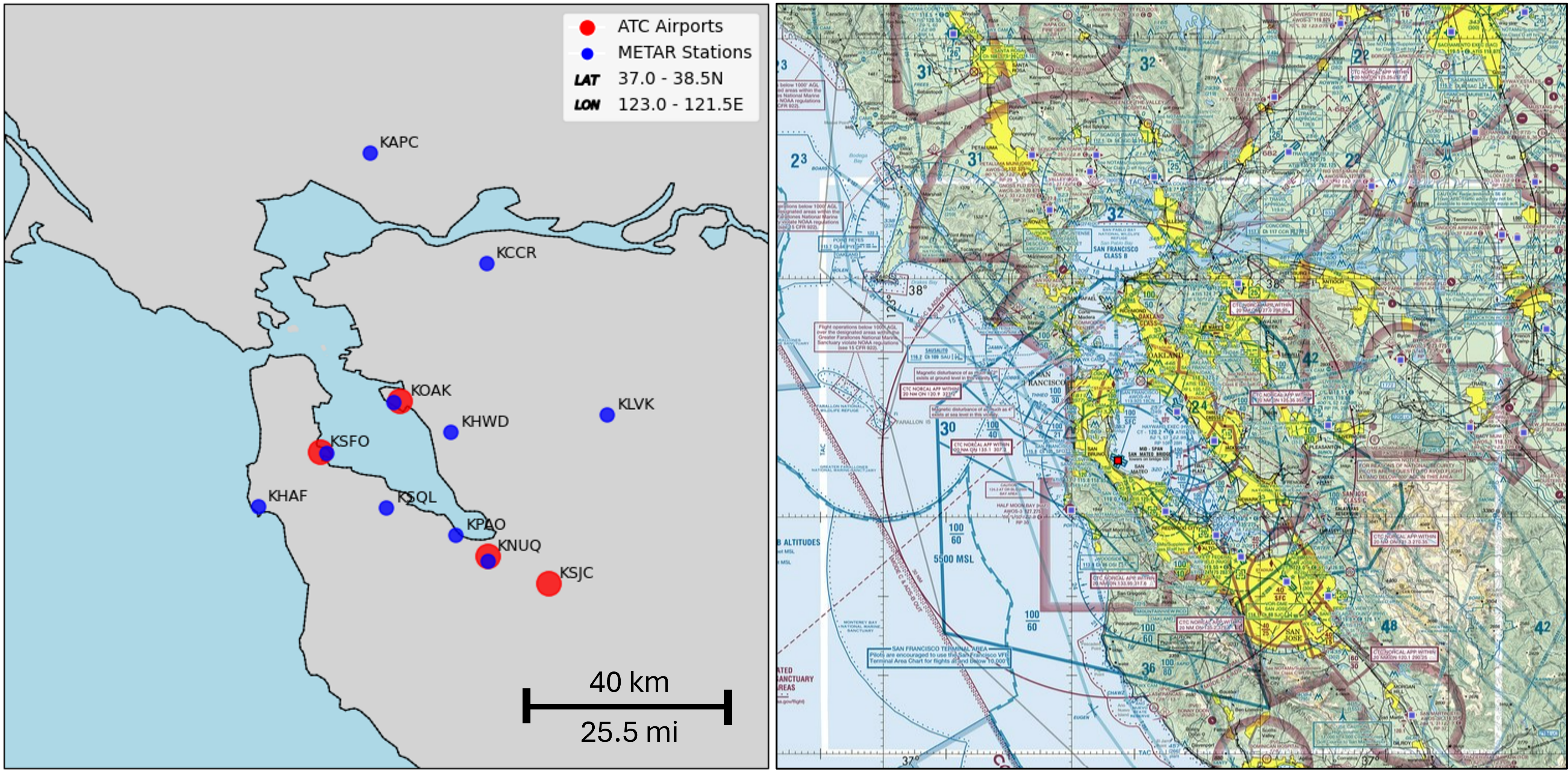}\\[2pt]
\makebox[0.5\textwidth]{\small a)}\makebox[0.5\textwidth]{\small b)}
\caption{Geographic scope of the data collection over the San Francisco Bay Area.
a) Map boundaries corresponding to the latitude-longitude bounding box of the bay,
which is also the geographical window used to scrape ADS-B information from the
OpenSky Network (\href{https://opensky-network.org}{opensky-network.org}). Red
circles mark the four airports whose ATC tower frequencies are streamed through
\href{https://www.liveatc.net}{LiveATC.net}. Blue markers indicate the weather
reporting stations from which METAR observations are periodically collected via
\href{https://aviationweather.gov}{aviationweather.gov}.
b) VFR (Visual Flight Rules) sectional chart of the same area, taken from
\href{https://vfrmap.com/}{vfrmap.com}, showing how densely concentrated the
airspace is, with overlapping controlled airspace boundaries, airways and
reporting points packed into a small region.}
\label{fig:airports_stations}
\end{figure}

We instantiate V2TATC over the San Francisco Bay Area as a case study for the geographic area shown in Fig.~\ref{fig:airports_stations}. The Bay Area is a particularly relevant region of interest for four reasons. (i) It hosts several major commercial airports, San Francisco International (KSFO), Oakland International (KOAK), and San Jose International (KSJC), as well as Moffett Field (KNUQ), all operating under the overlapping sectors of the Northern California TRACON. (ii) Traffic density is among the highest in the NAS, with a steady stream of arrivals and departures across the day. (iii) Commercial traffic and general aviation coexist on the same frequencies, so the data naturally captures both categories. (iv) The same airspace is expected to host some of the first operational trials of urban air mobility aircraft (e.g., \textit{electric vertical take-off and landing} (eVTOL) aircraft), which will further stress controller workload in the near future \cite{kam2025operational}. This concentration of features makes it possible to collect a rich multimodal dataset while keeping the geographic boundaries small. We describe how to collect synchronized voice and surveillance data over this airspace, how to clean and link them at scale, and how to train each component of the framework. We then present results on cross-modal retrieval, short-horizon trajectory forecasting, voice-conditioned trajectory generation, and the structure of the learned latent space.

A side product of the present work is the paired voice--trajectory dataset itself. To the best of our knowledge, no such dataset has been released publicly for the Bay Area, since existing ATC datasets are either voice-only or trajectory-only, and the few works that combine them do so on closed proprietary data or over smaller, uncongested regions. Building the dataset required audio-archive ingestion, real-time ADS-B polling, transcription, callsign extraction and temporal linking, all described in Section~\ref{sec:implementation}. We release a subset of our dataset\footnote{Available at \href{https://huggingface.co/datasets/Lbrusset/SF_bay_voice2traj_dataset}{\texttt{huggingface.co/datasets/Lbrusset/SF\_bay\_voice2traj\_dataset}}.} and the entire processing pipeline so that future work, whether on speech recognition, trajectory modeling, joint embeddings, or operational decision support tools, can build on it. The contributions of the present article are threefold. First, we formulate a learnable correspondence between voice and trajectory submanifolds and propose a framework that combines self-supervised pre-training, contrastive alignment, and bijective lifting through normalizing flows. Second, we describe an end-to-end implementation of V2TATC using publicly available data sources and standard self-supervised techniques, and we release both the assembled dataset and the code. Third, we report quantitative results that show that the two modalities can be aligned with a contrastive objective in spite of the noisy nature of ATC audio, and that the alignment supports useful downstream operations such as short-horizon forecasting and cross-modal retrieval.

\subsection*{Related Work}

Several articles have explored the use of machine learning in ATC, focusing primarily on a single ATC modality. \textit{Automatic Speech Recognition (ASR)} for ATC has progressed steadily, from constrained-grammar systems built on hidden Markov models~\cite{Schaefer2001} to transformer-based models trained on aviation-specific corpora such as ATCO2~\cite{zuluaga2022atco2}, on top of which downstream understanding tasks have been built~\cite{zuluaga2022bertraffic}. Trajectory modeling has likewise matured, from recurrent architectures~\cite{zhao2019lstm} to attention-based encoders operating on richer trajectory representations~\cite{guo2023flightbert}. Related works on \textit{large language models} (LLM) for ATC have shown that short-horizon trajectory forecasting is feasible with these transformer-based architectures~\cite{darrell2026representing}, and that agentic ATC can be performed from communication data in natural language alone~\cite{casanova2026air}. Readback error detection, which compares a transcribed pilot acknowledgment against the controller instruction~\cite{Helmke2022}, has a similar single-task focus. SIA-FTP~\cite{siaftp2023} uses spoken instructions as side information to improve flight-trajectory prediction, in a setup that is end-to-end supervised on the prediction task. V2TATC differs in that it learns a self-supervised joint embedding and then evaluates several downstream tasks on top of it. To our knowledge no prior work aligned raw ATC voice with surveillance trajectories at the embedding level. This article is to our knowledge the first step in this direction, and is conceptually related to the joint-embedding predictive architectures presented by LeCun~\cite{lecun2022jepa}.

\section{Problem Formulation}
\label{sec:problem}

\subsection{Situational Awareness in Congested Airspaces}
Situational awareness, in the sense of Endsley~\cite{endsley1995sa}, is the controller's ability to perceive, comprehend, and project the state of the airspace. In a busy terminal area, this state is high-dimensional: tens of aircraft, each with a dynamic trace, an identity, and an active stream of verbal instructions. Modern controller working positions display trajectory information in great detail through radar plots and ADS-B integration. The radio channel, by contrast, is still largely consumed as raw audio. While ASR tools are beginning to be deployed, voice is rarely interpreted and aligned with the surveillance stream in standard operational settings. The intuitive feature \emph{``click on this aircraft and recover what is being said about it''} is not natively supported by existing DSTs, because the surveillance display and the radio frequency are not aligned in any explicit way. Yet this operation is what controllers do mentally, dozens of times per hour. The cognitive load it generates has been quantified in the human-factors literature~\cite{Helmke2016}, and it is widely recognized as one of the structural limits of controller throughput.

We formalize this operation as a mapping between two manifolds. Let $\mathcal{X}_v$ be the space of voice transmissions and $\mathcal{X}_t$ be the space of trajectory windows. Both spaces describe a portion of the same airspace, but with different observables. The space $\mathcal{X}_v$ carries semantic information about the controller's intent, while $\mathcal{X}_t$ carries the observed resulting trajectory. To bring the two observables to a comparable semantic level, the trajectory-side works on the differences $\Delta x_t = x_{t+1} - x_t$ between two consecutive timesteps rather than on absolute positions. With this choice, we expect that intent can determine a relative maneuver and, conversely, that a relative maneuver carries enough information to reveal the underlying intent. The two observables are linked by the physical aircraft, but the link is not directly accessible from the raw signals. Our goal is to learn a shared representation space $\mathcal{Z}$ together with two encoders such that a voice phrase and the trajectory of the addressed aircraft project to nearby points in $\mathcal{Z}$, while unrelated pairs project to distant points. With this property, a controller could find the trajectory associated with a given voice transmission by nearest-neighbor lookup in $\mathcal{Z}$, and vice versa. In effect, $\mathcal{Z}$ serves as a learned cartography of the airspace.

\subsection{Two-Tower Joint Embedding Approach}
\begin{figure}[tp]
\centering
\begin{tikzpicture}[
  x=1cm, y=1cm,
  data/.style={draw=#1, line width=0.9pt, fill=#1!10, rounded corners=3pt,
               minimum width=3.9cm, minimum height=1.05cm, align=center,
               font=\small\bfseries},
  data/.default=black,
  block/.style={draw=#1, line width=0.9pt, fill=#1!4, rounded corners=3pt,
                minimum width=3.9cm, minimum height=0.95cm, align=center, font=\small},
  block/.default=black,
  arr/.style={-{Stealth[length=2.6mm]}, line width=0.9pt, draw=#1},
  arr/.default=black,
  lbl/.style={font=\scriptsize, inner sep=2pt},
  tower/.style={rounded corners=9pt, draw=#1!45, dashed, line width=0.7pt,
                fill=#1!3, inner sep=4.5mm},
  tower/.default=black,
]
 
\node (hs) at (-3.3,6.9) {\includegraphics[height=1.3cm]{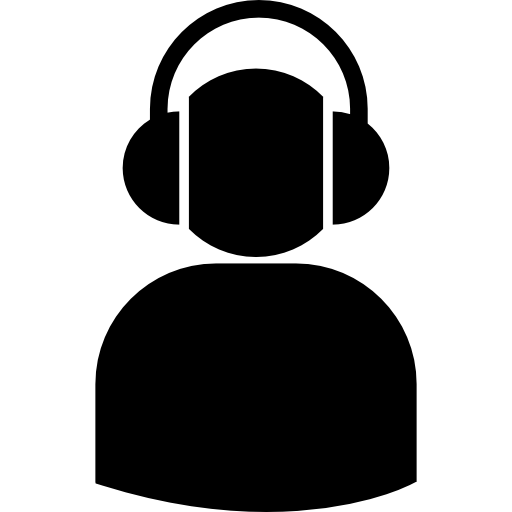}};
\node (ac) at ( 3.3,6.9) {\includegraphics[height=1.3cm]{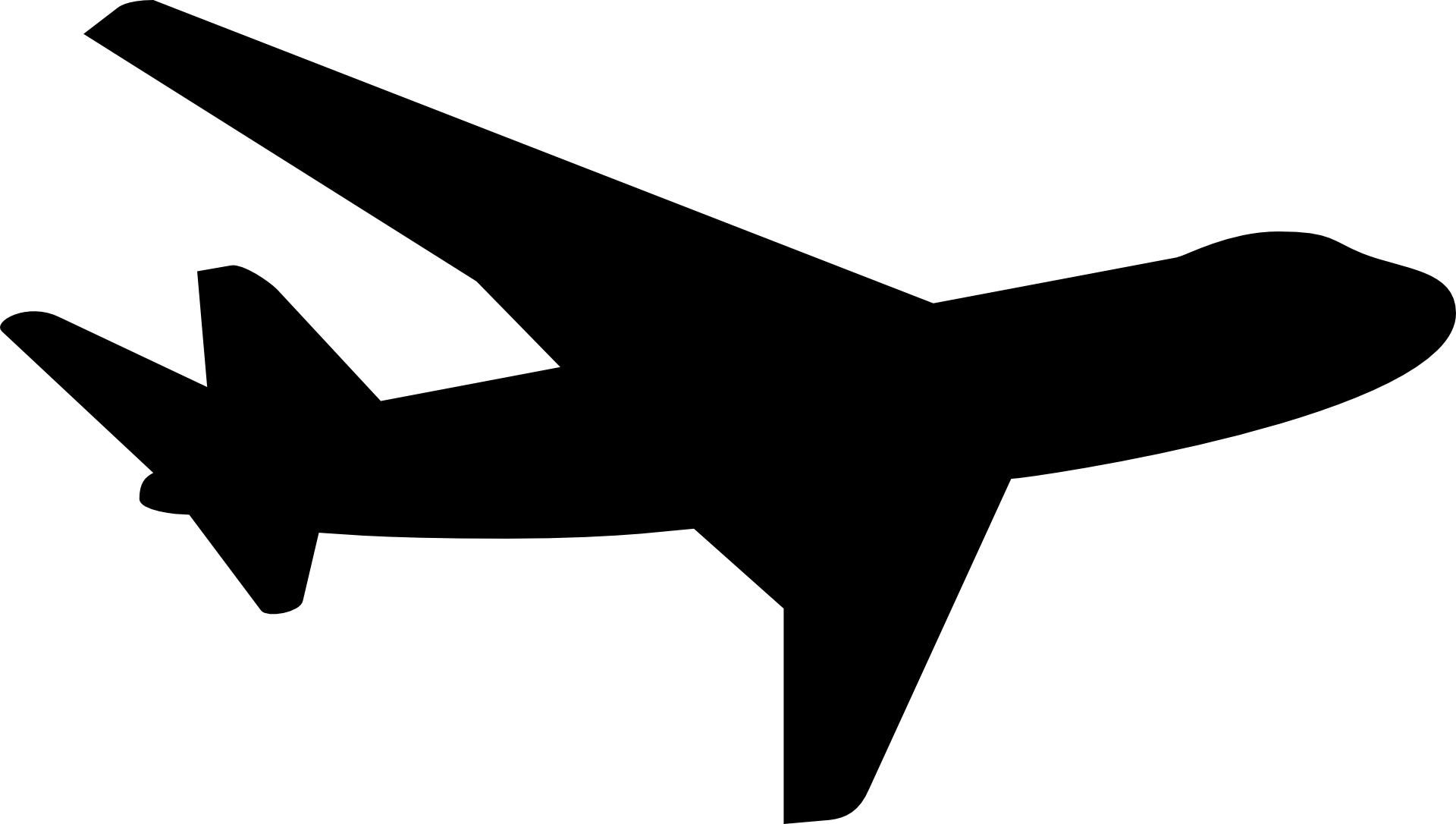}};
\node[lbl, text=voiceC, align=center, text width=4.6cm] (hsl) at (-3.3,7.95)
     {Controller and pilot\\radio exchange};
\node[lbl, text=trajC, align=center, text width=4.6cm] (acl) at (3.3,7.95)
     {Aircraft ADS-B\\surveillance track};
 
\node[data=voiceC]  (vd)   at (-3.3, 5.3) {Voice data ($\R^{B \times L}$)};
\node[block=voiceC] (vraw) at (-3.3, 3.8) {Voice encoder $\Ev$};
\node[block=voiceC] (vp)   at (-3.3, 2.3) {Projector $\Pv$};
 
\node[data=trajC]  (td)   at ( 3.3, 5.3) {Trajectory data ($\R^{B \times T \times F}$)};
\node[block=trajC] (traw) at ( 3.3, 3.8) {Trajectory encoder $\Et$};
\node[block=trajC] (tp)   at ( 3.3, 2.3) {Projector $\Pt$};
 
\node[draw=jointC, line width=1.0pt, fill=jointC!8, rounded corners=4pt,
      minimum width=9.2cm, minimum height=2.0cm] (joint) at (0,0.10) {};
\node[font=\small\bfseries, align=center] at (-2.6,0.10)
     {Joint space\\($\R^{B \times d_j}$)};
 
\begin{scope}[shift={(1.9,0.10)}]
  \foreach \x/\y in {-0.75/0.35, 0.35/-0.25, 1.45/0.30}{
    \draw[jointC!60, dashed, line width=0.5pt] (\x,\y) -- ++(0.42,-0.16);
    \draw[voiceC, fill=voiceC] plot[mark=*, mark size=1.7pt] coordinates {(\x,\y)};
    \draw[trajC, fill=trajC] plot[mark=triangle*, mark size=2.2pt]
          coordinates {(\x+0.42,\y-0.16)};
  }
  \node[lbl, align=center] at (0.55,-0.72)
       {\textcolor{voiceC}{$\bullet$}\,$\vproj$ \quad
        \textcolor{trajC}{$\blacktriangle$}\,$\tproj$};
\end{scope}
 
\draw[arr=voiceC] (hs) -- (vd);
\draw[arr=trajC]  (ac) -- (td);
\draw[arr=voiceC] (vd)   -- node[lbl, left,  xshift=-2pt] {$w$}      (vraw);
\draw[arr=trajC]  (td)   -- node[lbl, right, xshift=2pt]  {$X$}      (traw);
\draw[arr=voiceC] (vraw) -- node[lbl, left,  xshift=-2pt] {$\vemb$}  (vp);
\draw[arr=trajC]  (traw) -- node[lbl, right, xshift=2pt]  {$\tremb$} (tp);
\draw[arr=voiceC] (vp) -- ($(joint.north west)!0.30!(joint.north)$)
      node[lbl, pos=0.55, left, xshift=-2pt] {$\vproj$};
\draw[arr=trajC]  (tp) -- ($(joint.north east)!0.30!(joint.north)$)
      node[lbl, pos=0.55, right, xshift=2pt] {$\tproj$};
 
\node[draw=jointC, fill=jointC!10, rounded corners=3pt, inner sep=3pt,
      font=\small] (nce) at (0,2.3) {$\Lloss_\text{NCE}$};
\draw[{Stealth[length=2mm]}-{Stealth[length=2mm]}, jointC!70, dashed,
      line width=0.7pt] (vp.east) -- (nce.west);
\draw[{Stealth[length=2mm]}-{Stealth[length=2mm]}, jointC!70, dashed,
      line width=0.7pt] (nce.east) -- (tp.west);
 
\begin{scope}[on background layer]
  \node[tower=voiceC, fit=(hsl)(hs)(vd)(vraw)(vp)] {};
  \node[tower=trajC,  fit=(acl)(ac)(td)(traw)(tp)] {};
\end{scope}
 
\end{tikzpicture}
\caption{Two-tower joint embedding overview. The left tower processes voice, the
right tower processes trajectories, and both towers map their input to a common
space in which a contrastive objective pulls matching pairs together and pushes
mismatched pairs apart. $d_j$ is the dimension of the joint embedded space.}
\label{fig:two_tower}
\end{figure}
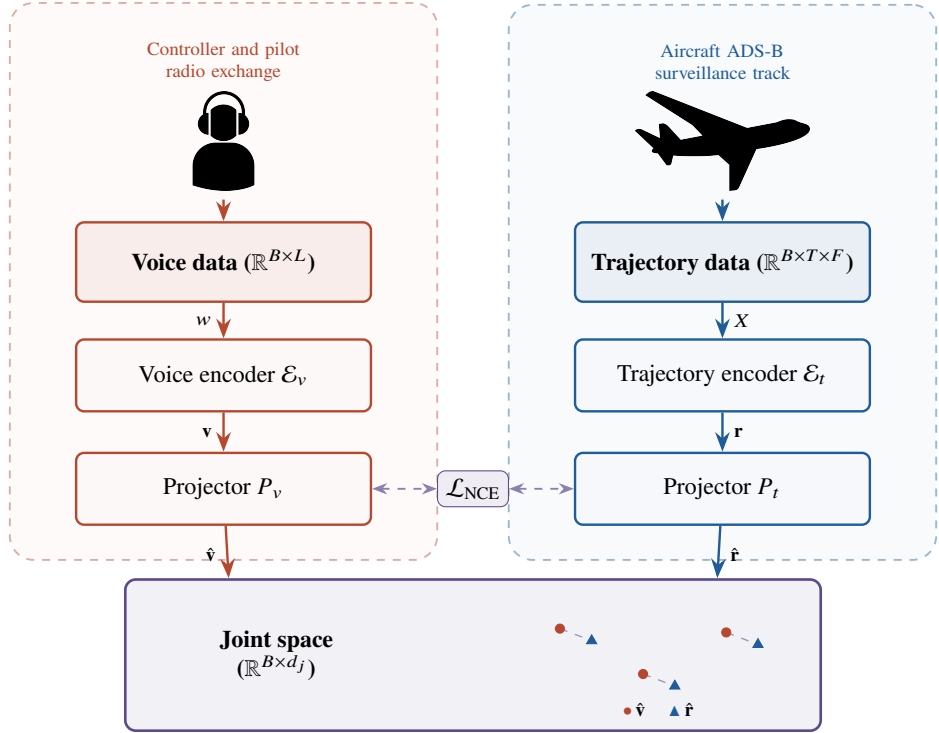

The proposed V2TATC approach follows the two-tower design that has become standard in multi-modal retrieval~\cite{radford2021clip,jia2021align}. One tower processes trajectory data, the other processes voice data. The two towers do not share parameters, and each is specialized for its own modality. They communicate through a single shared space at their output. Figure~\ref{fig:two_tower} shows the conceptual layout. The design is practical for several reasons. Candidate embeddings can be pre-computed offline, so online retrieval reduces to a single matrix-vector multiplication. Each tower also carries the right inductive bias for its own modality, with causal masking on the voice side and time-aware attention on the trajectory side. And because the towers are independent, either one can be swapped without retraining the other.

A contrastive objective is well suited here: it requires no ground-truth labels and no predefined classes, only a binary answer to the question of whether two observations describe the same aircraft at the same time. This binary signal is naturally available in our data, because every ADS-B trajectory carries the aircraft identity, \emph{International Civil Aviation Organization (ICAO)} 24-bit transponder code and callsign, and every transcribed voice phrase mentions a callsign. Pairing is therefore deterministic at the data level. The contrastive loss then turns this pairing into a metric in the joint space. Contrastive alignment alone, however, is not invertible. The projectors $\Pv$ and $\Pt$ are forward-only. Given an embedding they produce a point in $\mathcal{Z}$, but they cannot be run in reverse. To support the use case \emph{``given a point in $\mathcal{Z}$, recover the corresponding voice or trajectory embedding''}, we add a second stage that mimics a bijection between each modality space and $\mathcal{Z}$. This bijection is implemented as two normalizing flows~\cite{dinh2017realnvp}, one per tower. Once trained, the flows transform contrastive retrieval into a bidirectional translation between modalities, so that a voice can be turned into a synthetic trajectory representation, and vice versa.

\subsection{Bijectivity Beyond Retrieval}

Retrieval is the simplest application of a joint space, but it is not the most useful one operationally. A ATC DST may need to answer questions of the form ``if this voice instruction is followed, what would the trajectory look like?'' or ``what would a controller most plausibly say about this maneuver?'' Both questions require sampling in modality space, conditioned on a point in $\mathcal{Z}$. A contrastive projector cannot do this because its inverse is not defined. We use the term \emph{bijection} in a loose, geometric sense: not the strict set-theoretic bijection between $\mathcal{X}_v$ and $\mathcal{X}_t$ (which does not exist in practice, since a single trajectory can be described by many equivalent voice phrases and a single phrase can match several plausible trajectories), but a learned correspondence between the two low-dimensional submanifolds that voice and trajectory embeddings populate inside the joint space. The term \emph{bijectivity} also reflects our use of the RealNVP~\cite{dinh2017realnvp} network, whose built-in inversion maps one probability density to another and back through the same set of weights, so that a single network can be queried as voice~$\to$~trajectory or trajectory~$\to$~voice without retraining and without a separate inverse model.

\section{Approach}
\label{sec:approach}
This section gives the formal description of each block of V2TATC, and we keep the discussion conceptual. The concrete data sources, models, and hyperparameters are deferred to Section~\ref{sec:implementation}.

\subsection{Trajectory Tower}
A trajectory window $\Xtraj \in \R^{T \times F}$ is a fixed-length sequence of ADS-B samples observed at irregular times. Each timestep $x_t$ contains positions, velocity, heading, and vertical rate. The trajectory tower learns an encoder $\Et: \R^{T \times F} \to \R^{d_t}$ that maps such a window to a single vector or a sequence of token vectors. The encoder is trained in a self-supervised manner using a masked reconstruction objective. The MAE framework, introduced for vision by He~et~al.~\cite{he2022mae} and rooted in the masked language modeling tradition~\cite{devlin2019bert}, is well suited to trajectory data for two reasons. First, the supervision signal is the data itself, so no labels are required. This matters because trajectory datasets at scale come without semantic annotation. Second, the reconstruction task can only be solved if the encoder produces a globally consistent representation of the visible context. The resulting representation transfers well to downstream tasks~\cite{he2022mae}. This is also why we deliberately keep the decoder $g$ as small as possible: with little capacity of its own, the decoder cannot compensate for a poor encoder output, so it acts as a capacity bottleneck that pushes the burden of producing a globally consistent representation onto $\Et$. The decoder must nonetheless remain expressive enough to reconstruct the target signal, so that gradients flow back and effectively optimize the encoder parameters.

Let $\Mset \subset \{1,\dots,T\}$ be a random subset of timesteps, with $|\Mset| = \lfloor m T \rfloor$ where $m \in (0,1)$ is the masking ratio. The masked timesteps are replaced by a learned mask token $\mathbf{e}_{\text{mask}} \in \R^d$. The encoder processes the full corrupted sequence and produces hidden states $H = (h_1,\dots,h_T)$, and a lightweight decoder $g$ reconstructs the target representation $y_t$ on the masked positions. The training loss is
\begin{equation}
\label{eq:mae_loss}
\Lloss_{\text{MAE}} \;=\; \frac{1}{|\Mset| \cdot F} \sum_{t \in \Mset} \bigl\| g(h_t) - y_t \bigr\|_2^2 .
\end{equation}
The choice of target $y_t$ has a strong influence on what the encoder learns. We use the normalized first-order difference (delta $z$-score) of each feature: $y_t = (z_t - z_{t-1}) / \sigma_\delta$ where $z_t = (x_t - \mu) / \sigma$, $\mu, \sigma$ are per-feature mean and standard deviation, and $\sigma_\delta$ is the standard deviation of consecutive differences. This separation of scales prevents features at very different absolute ranges (e.g., longitude in km vs.\ vertical rate in m/s) from dominating the loss.

ADS-B samples are not produced at a strictly regular rate. Ground stations relay messages opportunistically, and missed reports lead to variable $\Delta t$ between consecutive points. To expose this irregular sampling to the encoder, we add a Time2Vec encoding~\cite{kazemi2019time2vec} of $\Delta t$ to each token. Time2Vec is a learned vector representation defined by
\begin{equation}
\label{eq:t2v}
\text{Time2Vec}(\tau)_k \;=\;
\begin{cases}
  \omega_0 \tau + \phi_0  & k = 0, \\
  \sin(\omega_k \tau + \phi_k)  & 1 \le k \le d_\tau - 1,
\end{cases}
\end{equation}
where $\omega_k, \phi_k$ are learnable parameters. This encoding combines a linear component (which captures monotonic time) and Fourier components (which capture periodic structure), without requiring a hand-designed positional scheme.

After pre-training, the encoder output is used as the trajectory embedding $\tremb \in \R^{d_t}$. We obtain a fixed-size vector by flattening the encoder hidden states across all $T$ timesteps. This choice preserves the temporal structure compared to mean pooling and makes more information available to the projector.

\subsection{Voice Tower}
A voice transmission $\wav$ is a variable-length waveform. The voice tower uses a frozen large-scale \emph{Automatic Speech Recognition (ASR)} encoder $\Ev$ that has been pre-trained on hundreds of thousands of hours of speech~\cite{radford2023whisper,baevski2020wav2vec2}. The encoder maps the waveform to a sequence of hidden states over audio frames, $\Ev(\wav) \in \R^{L' \times d_v}$, where $L'$ depends on the waveform length, and we apply a temporal mean pool to obtain a single embedding:
\begin{equation}
\label{eq:voice_pool}
\vemb \;=\; \frac{1}{L'} \sum_{\ell=1}^{L'} \Ev(\wav)_\ell \;\in\; \R^{d_v}.
\end{equation}

We keep $\Ev$ frozen for three reasons. First, the embedding it produces already encodes the semantic content of speech robustly across acoustic conditions, including noisy ones~\cite{radford2023whisper}. Second, fine-tuning a 1.5B-parameter speech encoder on ATC data risks catastrophic forgetting and demands much more training data than is currently available. Third, freezing $\Ev$ keeps the number of trainable parameters of the framework small, which makes contrastive training data-efficient.

The choice of pooling deserves a comment. Whisper produces a dense temporal sequence of frame-level hidden states: at roughly $100$ frames per second, a typical \SI{15}{\second} ATC clip yields close to $1{,}500$ frame embeddings of dimension $d_v = 1280$. At single-precision floating-point ($\SI{4}{\byte}$ per scalar), one clip alone takes about $1{,}500 \times 1280 \times \SI{4}{\byte} \approx \SI{7.7}{\mega\byte}$. Across the entire paired dataset of about $83{,}000$ voice samples, storing and processing the full frame-level Whisper output would therefore require roughly $\SI{650}{\giga\byte}$ of disk and RAM, which is impractical for the rest of the pipeline. Reducing each clip to a single vector through mean pooling brings the per-sample footprint down to $1280 \times \SI{4}{\byte} \approx \SI{5}{\kilo\byte}$ and the full corpus to under $\SI{500}{\mega\byte}$, three orders of magnitude lighter.

We selected mean pooling over alternatives such as max pooling because the arithmetic mean is a linear, information-preserving aggregator. Every frame contributes to the result, and no salient component of the embedding is discarded as it would be by an argmax. Mean pooling has also been reported as a strong baseline for non-autoregressive uses of pre-trained speech encoders~\cite{chen2022wavlm}. A consequence of any temporal pooling is that the mapping from the raw waveform to the pooled embedding is not bijective, since the pooled vector cannot be inverted back to the original frame sequence, which creates a strict information bottleneck between the voice tower and the joint space. The bijective stage of V2TATC operates downstream of this bottleneck, so it cannot recover the raw audio from a point in the joint space. In practice, this can be partially circumvented at inference time using a look-up table of the training data. Given a query point in the joint space, the nearest training samples can be retrieved and their associated audio re-used directly. The pooled embedding therefore supports retrieval and cross-modal generation in latent space, while access to a faithful waveform is delegated to the training data itself.

\subsection{Joint Embedding via Contrastive Alignment}
\label{sec:approach_contrastive}
The two-tower outputs live in different spaces ($\R^{d_v}$ and $\R^{d_t}$, with $d_v \ne d_t$ in general). We project both into a common space $\R^{d_j}$ through two \textit{Feed-Forward Neural Network} (FFNN) projectors:
\begin{equation}
\label{eq:projectors}
\vproj = \Pv(\vemb), \qquad \tproj = \Pt(\tremb), \qquad \vproj, \tproj \in \R^{d_j} .
\end{equation}
We then $\ell_2$-normalize both projections: $\vnorm = \vproj / \|\vproj\|_2$ and $\tnorm = \tproj / \|\tproj\|_2$. The normalization moves the embeddings to the unit hypersphere, which has the effect of bounding inner products to $[-1, 1]$ and turning them into cosine similarities. We then train $(\Pv, \Pt)$ with the symmetric \emph{Information Noise-Contrastive Estimation} (InfoNCE) objective~\cite{oord2018cpc,radford2021clip}. For a batch of $B$ paired observations, define the similarity matrix
\begin{equation}
\label{eq:sim}
S_{ij} \;=\; \vnorm_i^\top \tnorm_j, \qquad i, j \in \{1, \dots, B\}.
\end{equation}
The InfoNCE loss treats each pair $(i, i)$ as its own positive and all other in-batch pairs as negatives, in both directions:
\begin{equation}
\label{eq:infonce}
\Lloss_{\text{NCE}} \;=\; -\frac{1}{2B} \sum_{i=1}^B \left[ \log \frac{e^{S_{ii}/\tau}}{\sum_{j=1}^B e^{S_{ij}/\tau}} + \log \frac{e^{S_{ii}/\tau}}{\sum_{j=1}^B e^{S_{ji}/\tau}} \right] ,
\end{equation}
where $\tau > 0$ is a learnable temperature.

The temperature $\tau$ controls the sharpness of the softmax distribution over candidates. A small $\tau$ amplifies similarity differences and produces a sharper distribution, which favors precise discrimination between positive and hard negative pairs. A large $\tau$ softens the distribution and encourages a more uniform assignment. In CLIP, $\tau$ is typically learned in log-space ($\tau = \exp(\theta)$ with $\theta$ trainable) and clamped to a positive interval to prevent collapse to zero~\cite{radford2021clip}. We adopt the same parameterization. Theoretical analyses of InfoNCE link a small $\tau$ to a tighter lower bound on mutual information between the two modalities~\cite{poole2019mibounds}, which provides a principled justification for the empirical observation that $\tau$ tends to drift downward during training. Earlier multi-modal alignment work often relied on triplet or margin losses, in which one anchor, one positive, and one negative are sampled per training step. InfoNCE generalizes this by using all in-batch examples as negatives simultaneously, which improves data efficiency and gradient quality at large batch sizes~\cite{chen2020simclr}. The cost is that very large batches are required to expose the model to enough hard negatives, which is why batch size is treated as a first-class hyperparameter in CLIP-style training.

\subsection{Bijective Lifting via Normalizing Flows}
\label{sec:approach_bijection}
After contrastive training, voice and trajectory embeddings that describe the same aircraft are close in the joint space. However, the projectors $\Pv$ and $\Pt$ are not invertible, because they are general FFNNs whose inverses are not defined in closed form. To enable bidirectional translation between modalities, we train two normalizing flows $\fv, \ft : \R^{d_j} \to \R^{d_j}$ that mimic invertible maps between each modality space (padded to $\R^{d_j}$) and the joint space. A normalizing flow is a sequence of invertible parametric transformations with tractable Jacobian determinant. The change-of-variables formula then yields an exact and differentiable expression for the log-likelihood under the transformed distribution~\cite{rezende2015flows}. The original \emph{Real-valued Non-Volume Preserving} (RealNVP) construction~\cite{dinh2017realnvp} introduced the affine coupling layer, in which the input is split in two halves $(\mathbf{x}_1, \mathbf{x}_2)$ along a binary mask $\mathbf{m} \in \{0,1\}^{d_j}$ and transformed as
\begin{equation}
\label{eq:coupling}
\mathbf{y}_1 = \mathbf{x}_1, \qquad \mathbf{y}_2 = \mathbf{x}_2 \odot \exp\bigl( s(\mathbf{x}_1) \bigr) + t(\mathbf{x}_1),
\end{equation}
where $s$ and $t$ are neural-network conditioners and $\odot$ is the element-wise product. The inverse is available in closed form:
\begin{equation}
\label{eq:coupling_inv}
\mathbf{x}_2 = \bigl( \mathbf{y}_2 - t(\mathbf{x}_1) \bigr) \odot \exp\bigl( -s(\mathbf{x}_1) \bigr), \qquad \mathbf{x}_1 = \mathbf{y}_1 .
\end{equation}
The Jacobian of the coupling is triangular, so its log-determinant reduces to $\log|\det J| = \sum_k s(\mathbf{x}_1)_k$. Stacking coupling layers with alternating masks transforms all dimensions progressively. ActNorm layers~\cite{kingma2018glow}, introduced in Glow, are inserted between coupling blocks to normalize activations per dimension, which improves conditioning and training stability. The structure of an affine coupling layer is illustrated in Fig.~\ref{fig:coupling_diagram}. More detail on the overall structure of RealNVP is given in Appendix~D, Fig.~\ref{fig:realnvp_explained}.

\begin{figure}[tp]
\centering
\includegraphics[width=.55\textwidth]{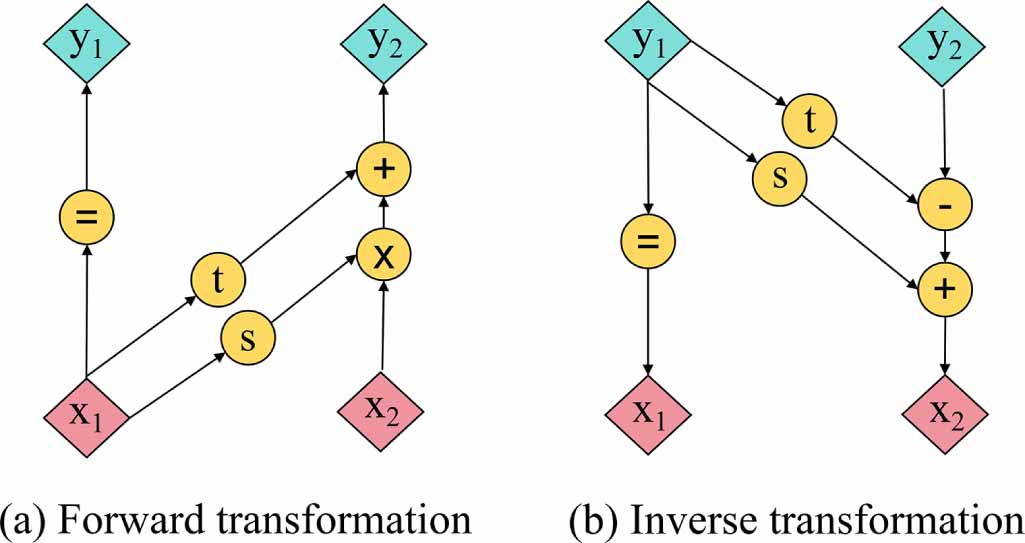}
\caption{Affine coupling layer of a RealNVP flow~\cite{dinh2017realnvp}. Panel (a) shows the forward transformation, in which one half of the input is passed through unchanged and used to condition the scale $s$ and translation $t$ applied to the other half. Panel (b) shows the inverse transformation, obtained analytically by reversing the affine step on the second half while reusing the same conditioning input. The block is invertible by construction and has a triangular Jacobian.}
\label{fig:coupling_diagram}
\end{figure}

The voice space has dimension $d_v = 1280$, the trajectory space has dimension $d_t = 1792$, and the joint space lives in $\R^{d_j} = \R^{1024}$. Because a normalizing flow preserves dimensionality, the bijection must operate at a single common width. We set this working width to $D_\text{max} = \max(d_v, d_t, d_j) = 1792$, and zero-pad whichever vector is smaller (typically the voice embedding and the joint vector) up to $D_\text{max}$ before passing it through the flow. Rather than being discarded, the padded dimensions are retained in the reconstruction loss, and the flow learns to map them onto near-zero values. The resulting transformation is a distribution-to-distribution bijection in which the surplus dimensions are contracted toward~$0$, which is precisely what lets a single equal-width invertible network bridge two embedding spaces of different intrinsic dimensionality.

\subsection{Training objective}
The two flows are trained jointly with a triple loss that combines forward, backward, and cross-modal consistency:
\begin{equation}
\label{eq:nvp_loss}
\Lloss \;=\; \lambda_f \sum_{m \in \{v,t\}} \!\! \bigl\| f_m(\mathbf{e}_m) - \jemb_m \bigr\|_2^2
  + \lambda_b \sum_{m \in \{v,t\}} \bigl\| f_m^{-1}(\jemb_m) - \mathbf{e}_m \bigr\|_2^2
  + \lambda_c \bigl\| \fv(\mathbf{e}_v) - \ft(\mathbf{e}_t) \bigr\|_2^2 ,
\end{equation}
where $\jemb_m$ is the joint-space target produced by the frozen contrastive projector. To train the higher-capacity RealNVP without overfitting and to make it faithfully reproduce the frozen FFNN projector over its whole domain, we augment the training set with synthetic pairs. A mesh of points sampled from the encoder-output representation space is passed through the frozen FFNN, which yields a much larger set of (sample,~target) pairs. The cross term, by contrast, is still trained only on the real (non-synthetic) paired samples, so as to preserve an optimal cross-modal alignment. The forward term forces each flow to match the contrastive projector. The backward term forces the inverse to recover the original embedding, which is the key property for cross-modal generation. The cross term aligns the two flows so that paired voice and trajectory map to the same joint point.

Note that $\jemb_m$ is a frozen target, so no gradient flows back into the upstream encoders or into the contrastive projectors that produced it. V2TATC's framework is therefore trained stagewise: the trajectory encoder $\Et$ is pre-trained first; the contrastive projectors $(\Pv, \Pt)$ are trained next on top of frozen $(\Ev, \Et)$; and finally the two flows $(\fv, \ft)$ are trained on top of frozen $(\Ev, \Et, \Pv, \Pt)$. This staged setup is a deliberate choice. Jointly fine-tuning all stages would compound the computational cost of each upstream model, and the limited size of the paired dataset makes such end-to-end training unstable in practice. Stagewise training also makes it easier to swap any single component without invalidating the others. Once trained, the chain $\fv$ then $\ft^{-1}$ implements a \textit{voice $\rightarrow$ trajectory} translation in latent space, and the symmetric chain implements the reverse direction. The result is a unified architecture in which the four operations \emph{voice~$\to$~joint}, \emph{joint~$\to$~voice}, \emph{trajectory~$\to$~joint}, \emph{joint~$\to$~trajectory} are all available through forward or inverse passes of the same two networks.

\begin{figure}[tp]
\centering
\includegraphics[width=\textwidth]{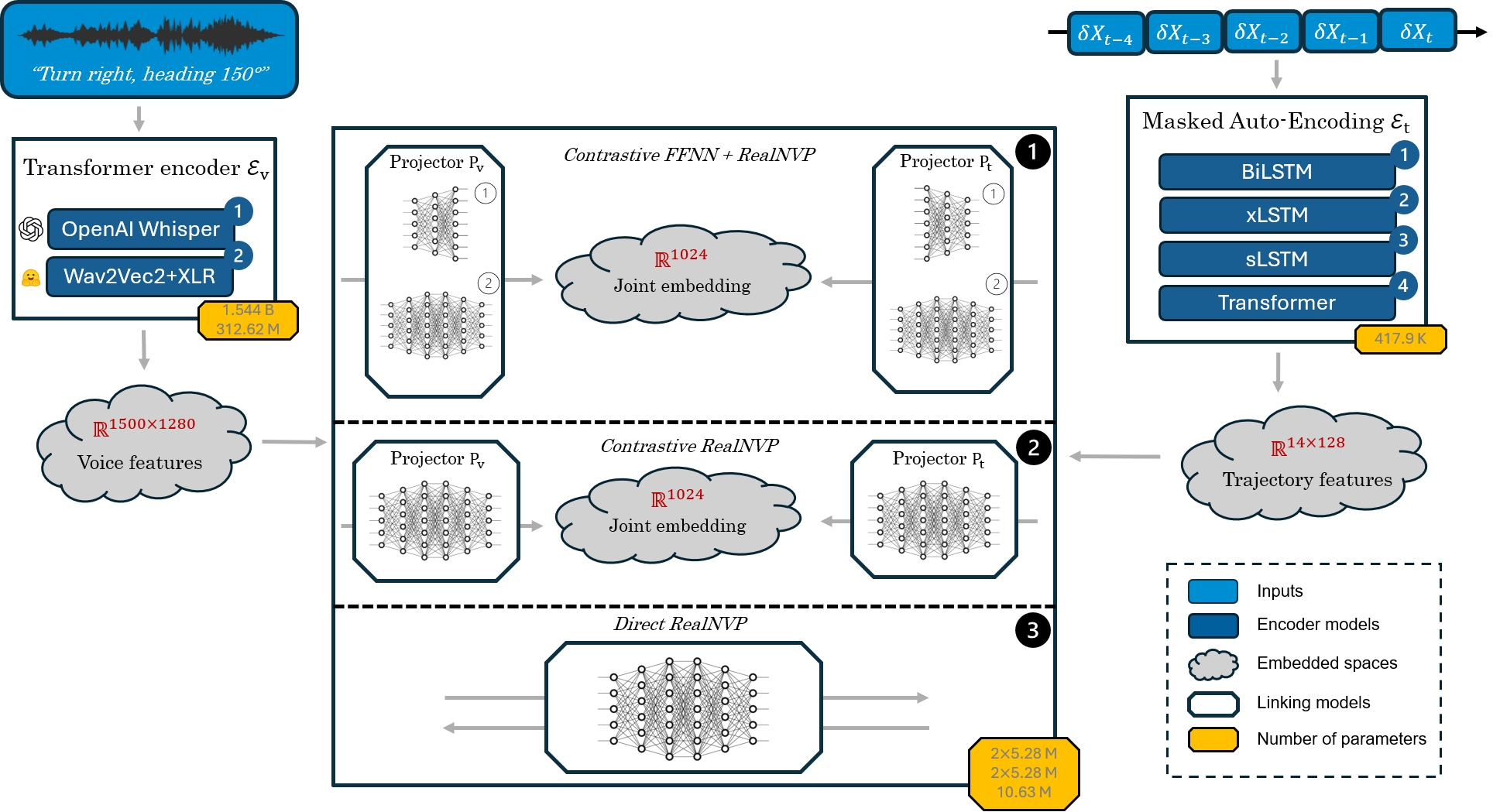}
\caption{Overall architecture and the three alignment variants. Boxes numbered in the figure are the alternatives benchmarked for each block. \textbf{Voice tower (left):} a frozen open-source ASR foundation model, either (1)~Whisper~\cite{radford2023whisper} or (2)~Wav2Vec~2.0~\cite{baevski2020wav2vec2}. Whisper is retained (Appendix~C). \textbf{Trajectory tower (right):} a MAE built around one of four sequence encoders trained from scratch on ADS-B deltas $\delta \Xtraj_t$, namely (1)~a bidirectional LSTM, (2)~a full xLSTM with matrix memory~\cite{beck2024xlstm}, (3)~a scalar xLSTM (sLSTM cells only), and (4)~a Transformer~\cite{vaswani2017attention}. The Transformer is retained (Appendix~B). \textbf{Joining stage (center):} three strategies compared in Section~\ref{sec:results_ablation}, namely (1)~contrastive FFNN projectors followed by a RealNVP bijective stage, which is the baseline used throughout the article, (2)~a single pair of RealNVP flows trained contrastively, doing both jobs at once, and (3)~one RealNVP mapping the two modality embeddings directly into one another, with no joint space, which isolates the contribution of the contrastive stage. Red vector spaces give the dimension of the space each sample lives in, and orange badges give the parameter count of each block. The voice tower connects to the joining stage through a temporal mean pool over the frame-level hidden states.}
\label{fig:arch_variations}
\end{figure}

\section{Implementation}
\label{sec:implementation}
This section describes a concrete instantiation of the framework over the San Francisco Bay Area. ADS-B trajectories are polled in real time by a multi-threaded orchestrator, each source being queried on its own schedule, while the radio audio covering the same period is obtained afterwards from the LiveATC archives. Covering both modalities over a single common time window is what makes the later temporal alignment possible, and the two subsections below detail each source in turn. The notation introduced in Section~\ref{sec:approach} is kept unchanged. An overview of all the architectural pieces that we tested and of the three alignment variants compared in Section~\ref{sec:results_ablation} is given in Fig.~\ref{fig:arch_variations}, and the rest of this section discusses each block one by one.

\subsection{Trajectory Data}

ADS-B is a surveillance system in which each aircraft determines its own state vector from onboard navigation sources and broadcasts it periodically, without interrogation from the ground. Each message carries the ICAO~24-bit transponder code, the callsign, a timestamp, \emph{World Geodetic System~1984} (WGS84) coordinates $(\phi, \lambda, h)$, ground speed $v$, true track $\theta$, and vertical rate $\dot{h}$. We collect ADS-B data from two public sources. The first one is the OpenSky Network~\cite{schafer2014opensky} (\href{https://opensky-network.org}{opensky-network.org}), a research-oriented feed that aggregates messages from a worldwide volunteer receiver network and offers a stable historical API. The second one is \href{https://adsb.lol}{adsb.lol}, an unauthenticated open feed that provides higher polling frequency at the cost of less complete historical coverage. For operational reasons we eventually retained only the OpenSky Network feed for the experiments reported here, but future work would benefit from incorporating the higher-frequency adsb.lol stream as well.

OpenSky state vectors are polled at \SI{30}{\second} intervals, and adsb.lol at roughly \SI{1}{\second} intervals when available. The geographic boundaries of the surveillance and voice streams are summarized in Fig.~\ref{fig:airports_stations}. ADS-B is collected over a rectangular bounding box around the Bay Area, and voice audio is captured at the four airports whose ATC frequencies are streamed by LiveATC. Table~\ref{tab:adsb_stats} summarizes the raw ADS-B data collected over the Bay Area, which serves as the pre-training material for the trajectory encoder.

\begin{table}[hbt!]
\caption{\label{tab:adsb_stats}Statistics of the raw ADS-B trajectory dataset collected from the OpenSky Network over the San Francisco Bay Area. The final sequences are used as the pre-training corpus for the trajectory encoder.}
\centering
\begin{tabular}{ll}
\toprule
\textbf{Property} & \textbf{OpenSky Network} \\
\midrule
Source area         & $37.0$--$38.5^\circ$\,N, $123.0$--$121.5^\circ$\,W \\
Raw ADS-B points    & 8{,}673{,}150 \\
Sequence length     & 14 timesteps ($\sim$\SI{6}{\minute}) \\
Window stride       & 1 \\
Final sequences     & 5{,}603{,}034 \\
Sampling frequency  & $\sim$\SI{26}{\second} (std $\sim$\SI{7}{\second}) \\
Vertical rate       & Transponder (barometric) \\
Split strategy      & By ICAO~24, callsign, timestamp \\
Train / Val / Test  & 85\% / 10\% / 5\% \\
\bottomrule
\end{tabular}
\end{table}

Raw geodetic coordinates are not suitable as a neural-network input. Three issues need to be fixed. First, longitude is non-linear, since \SI{1}{\degree} of longitude does not correspond to the same physical distance at every latitude. Second, the three spatial dimensions have incompatible units (degrees vs.\ meters), which would dominate any per-feature normalization. Third, true track is a circular variable ($359^\circ \to 0^\circ$ wrap-around), which standard $z$-score normalization cannot handle. We resolve these issues by converting each ADS-B sample into a local tangent-plane Cartesian frame centered on a reference point $(\phi_0, \lambda_0) = (37^\circ\,\text{N}, 123^\circ\,\text{W})$:
\begin{equation}
\label{eq:cart}
\begin{aligned}
x &= (R + h)(\lambda - \lambda_0)\cos\phi, \\
y &= (R + h)(\phi - \phi_0), \\
z &= h.
\end{aligned}
\end{equation}
where $R = \SI{6371}{\kilo\meter}$ is the mean Earth radius and $h$ is the geometric altitude of the aircraft above the reference ellipsoid. Velocity components are projected onto the same axes through $v_x = v \cos\theta$, $v_y = v \sin\theta$, $v_z = \dot{h}$. The resulting feature vector has $F = 6$ dimensions per timestep, all expressed in meters or meters per second. Each flight segment is then sliced into fixed-length windows $\Xtraj \in \R^{T \times F}$ of $T = 14$ consecutive samples. Window beginnings and ends are detected automatically when (i) the callsign changes or (ii) the gap between consecutive samples exceeds two minutes. A continuity buffer of 13 points is carried over between consecutive raw data files so that long flights are not truncated artificially at file boundaries.

The trajectory encoder $\Et$ is pre-trained as a MAE following the principle of Section~\ref{sec:approach}. The MAE was originally introduced for natural images~\cite{he2022mae}, and the recipe extends naturally to other domains as long as each input can be tokenized into a sequence. In our case, each ADS-B timestep is one token, and the encoder processes a sequence of $T = 14$ tokens. The decoder reconstructs the normalized delta $z$-score of each feature on the masked positions. We organize pre-training in two stages. Stage~1 operates directly on spherical coordinates $(\phi, \lambda, h, v, \theta, \dot{h})$ with normalized first-order differences as both encoder input and decoder target. Stage~2 operates on the Cartesian features $(x, y, z, v_x, v_y, v_z)$ of Eq.~\eqref{eq:cart}, with absolute normalized positions as input and Cartesian deltas as the reconstruction target. This eliminates the circular discontinuity of $\theta$ and the unit mismatch between angular and linear features.

\begin{figure}[htbp]
\centering
\includegraphics[width=\textwidth]{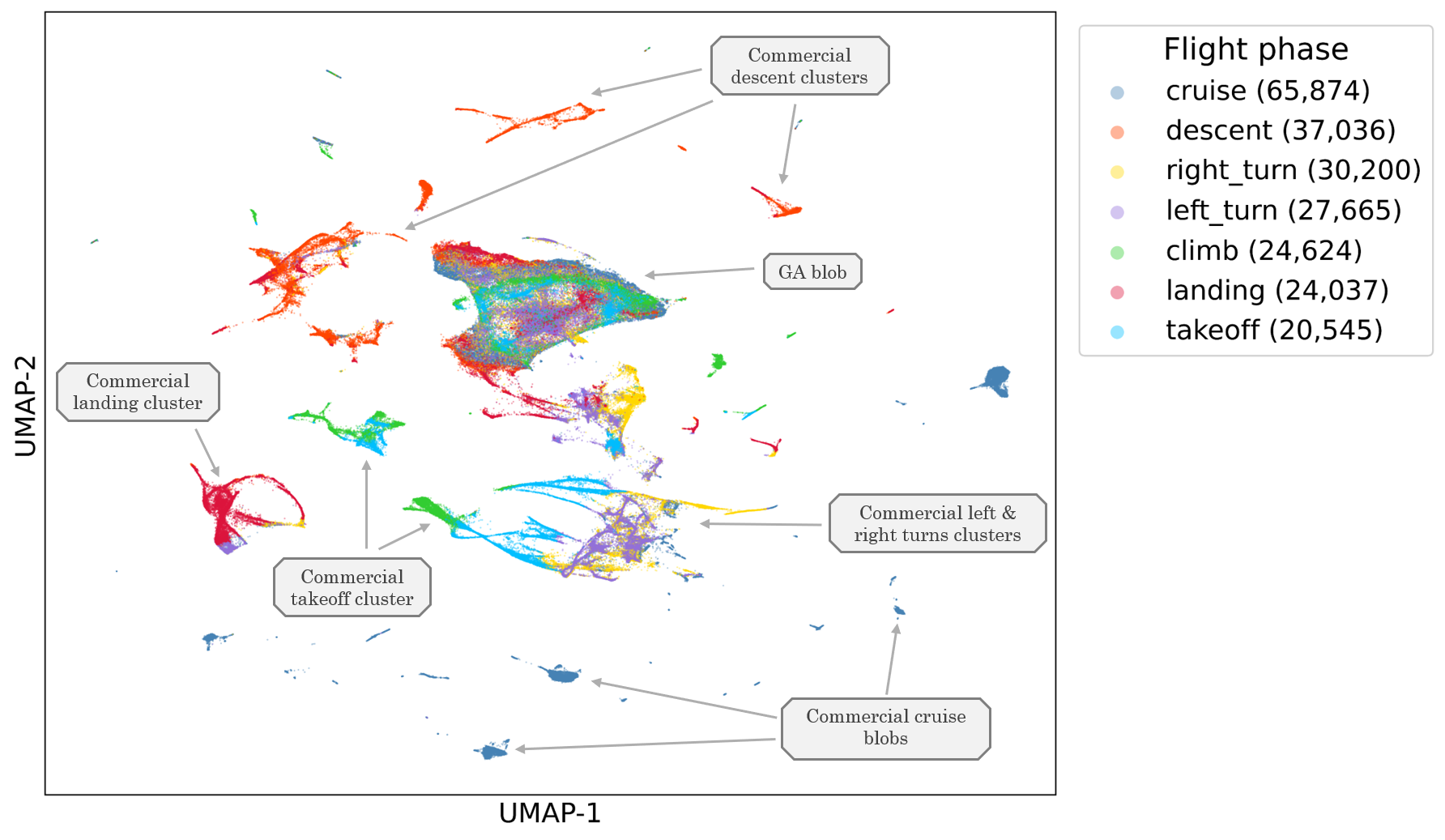}
\caption{UMAP projection of the trajectory latent space produced by the Transformer MAE on the OpenSky San Francisco dataset. The points organize by flight phase even though no phase label was used during pre-training, which indicates that the encoder captures the deepest essence of each window. Each cluster corresponds to an elementary behavior, namely a curved ground track or a change of altitude, and this alone is enough to tell general aviation apart from commercial traffic. Commercial flights follow standard procedures and repeat the same maneuvers from one flight to the next, so they form the tight and clearly separated clusters annotated on the plot. General aviation trajectories are far less standardized and collapse instead into a single continuous blob that connects all phases. Part of the structure is geographic rather than kinematic, since each Bay Area airport produces its own family of approach and departure trajectories and therefore its own cluster for a given phase label.}
\label{fig:transformer_proj}
\end{figure}

We compared four backbone architectures for $\Et$, all shown in the trajectory tower of Fig.~\ref{fig:arch_variations}, with hidden dimension $d = 128$ and four layers: a Transformer~\cite{vaswani2017attention}, a bidirectional LSTM, a scalar xLSTM (sLSTM cells only), and a full xLSTM with matrix memory~\cite{beck2024xlstm}. The Transformer is retained as the encoder for the rest of the framework. The detailed quantitative comparison of the four backbones is deferred to Appendix~B. The trajectory embedding $\tremb$ is obtained by flattening the encoder output across timesteps, giving $d_t = T \cdot d = 14 \cdot 128 = 1792$. Figure~\ref{fig:transformer_proj} shows the UMAP projection of the trajectory latent space produced by the Transformer MAE. This is the representation fed to the contrastive projector $\Pt$ in the next stage.

\subsection{Voice Data}

ATC radio audio is operated by \href{https://www.liveatc.net}{LiveATC.net}~\cite{liveatc}, a community-based network of volunteer receivers that broadcasts live ATC frequencies publicly and provides historical audio archives on request. For this study, we obtained the audio archives corresponding to eight Bay Area frequencies, listed in Table~\ref{tab:tower_counts}, spanning ground, tower, approach, and en-route sectors of KSFO, KOAK, KSJC, and Moffett Field. The archives cover \SI{12}{\hour} per day over four consecutive days, from \emph{11 to 14 February 2026}, and are delivered as a sequence of \SI{30}{\minute} MP3 chunks per frequency. ID3 metadata embedded in each chunk contains the exact UTC start and end times, which is essential for later alignment with the ADS-B stream. A small UTC-anchored buffer is kept between consecutive chunks to avoid losing transmissions that span the chunk boundary.

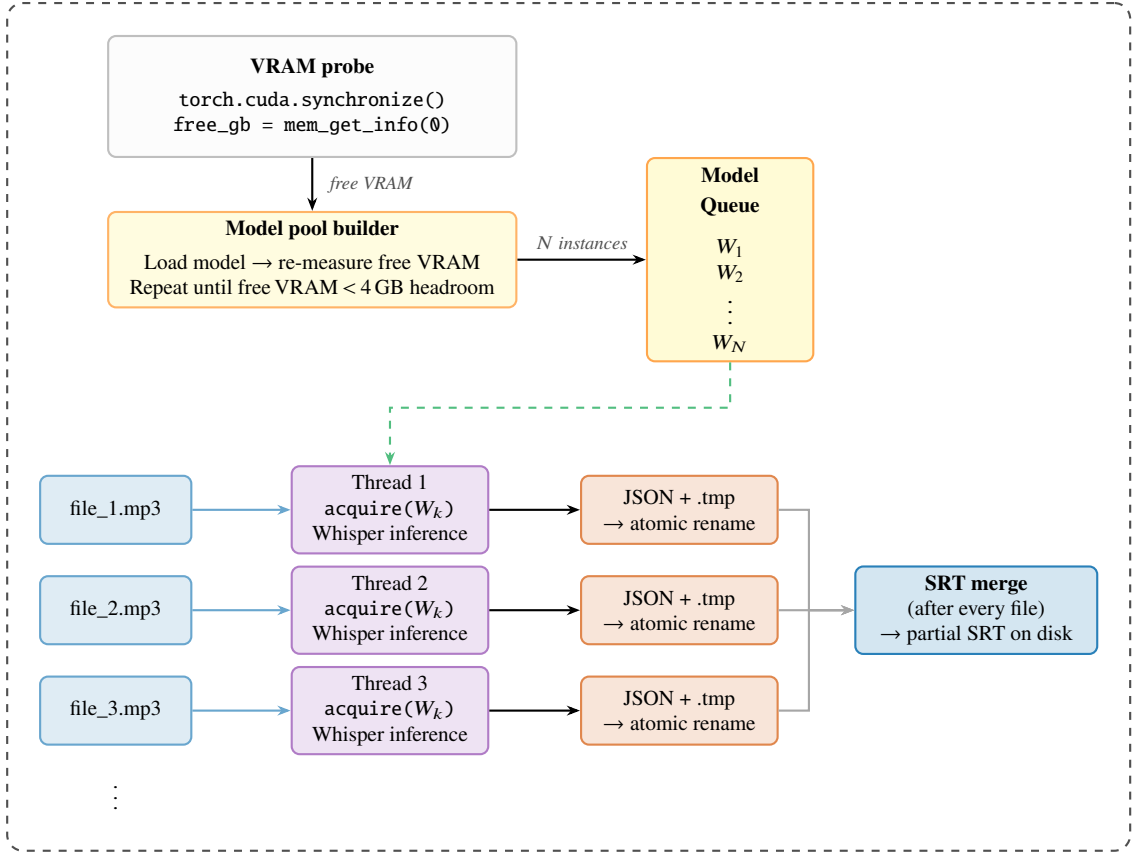
\begin{figure}[tp]
\centering
\begin{tikzpicture}[
  node distance = 0.7cm and 1.1cm,
  box/.style  = {draw, thick, rounded corners=3pt, align=center,
                 font=\footnotesize, minimum height=0.9cm},
  gpu/.style  = {box, fill=audiogreen!20, draw=audiogreen!80,
                 minimum width=2.4cm},
  mp3/.style  = {box, fill=adsbblue!15, draw=adsbblue!70,
                 minimum width=2.0cm},
  queue/.style= {box, fill=yellow!20, draw=orange!70,
                 minimum width=2.6cm},
  worker/.style={box, fill=linkpurple!15, draw=linkpurple!70,
                 minimum width=2.6cm, minimum height=1.1cm},
  outbox/.style = {box, fill=jointorange!15, draw=jointorange!70,
                 minimum width=2.6cm},
  arrow/.style= {-{Stealth[length=5pt]}, thick},
  label/.style= {font=\scriptsize\itshape, color=darkgray},
]

\node[box, fill=gray!2, draw=gray!50, minimum width=5.4cm, minimum height=1.6cm,
      align=center] (vramprobe) at (0,0)
  {\textbf{VRAM probe}\\[3pt]
   \texttt{torch.cuda.synchronize()}\\
   \texttt{free\_gb = mem\_get\_info(0)}};

\node[box, fill=yellow!15, draw=orange!60, minimum width=5.4cm,
      below=0.7cm of vramprobe, align=center] (poolbuilder)
  {\textbf{Model pool builder}\\[3pt]
   Load model $\to$ re-measure free VRAM\\
   Repeat until free\,VRAM\,$<$\,4\,GB headroom};

\draw[arrow] (vramprobe) -- (poolbuilder)
  node[midway, right=3pt, label] {free VRAM};

\node[queue, right=1.7cm of poolbuilder, minimum height=2.4cm,
      minimum width=2.2cm, align=center] (queue)
  {\textbf{Model}\\[2pt]\textbf{Queue}\\[6pt]
   $W_1$\\$W_2$\\$\vdots$\\$W_N$};

\draw[arrow] (poolbuilder) -- (queue)
  node[midway, above, label] {$N$ instances};

\node[mp3, below=2.2cm of poolbuilder, xshift=-2.6cm] (f1) {file\_1.mp3};
\node[mp3, below=0.4cm of f1]                         (f2) {file\_2.mp3};
\node[mp3, below=0.4cm of f2]                         (f3) {file\_3.mp3};
\node[font=\scriptsize, below=0.2cm of f3]            (fdots) {$\vdots$};

\node[worker, right=1.3cm of f1] (t1)
  {Thread 1\\{\footnotesize\texttt{acquire($W_k$)}}\\{\footnotesize Whisper inference}};
\node[worker, right=1.3cm of f2] (t2)
  {Thread 2\\{\footnotesize\texttt{acquire($W_k$)}}\\{\footnotesize Whisper inference}};
\node[worker, right=1.3cm of f3] (t3)
  {Thread 3\\{\footnotesize\texttt{acquire($W_k$)}}\\{\footnotesize Whisper inference}};

\draw[arrow, adsbblue!70] (f1) -- (t1);
\draw[arrow, adsbblue!70] (f2) -- (t2);
\draw[arrow, adsbblue!70] (f3) -- (t3);

\draw[arrow, audiogreen!80, dashed] (queue.south) -- ++(0, -0.6) -| (t1.north);

\node[outbox, right=1.2cm of t1] (j1) {JSON + .tmp\\$\to$ atomic rename};
\node[outbox, right=1.2cm of t2] (j2) {JSON + .tmp\\$\to$ atomic rename};
\node[outbox, right=1.2cm of t3] (j3) {JSON + .tmp\\$\to$ atomic rename};

\draw[arrow] (t1) -- (j1);
\draw[arrow] (t2) -- (j2);
\draw[arrow] (t3) -- (j3);

\node[box, fill=adsbblue!20, draw=adsbblue, minimum width=3.2cm,
      right=1.0cm of j2, yshift=-0.0cm] (srt)
  {\textbf{SRT merge}\\(after every file)\\$\to$ partial SRT on disk};

\draw[arrow, gray!70] (j1.east) -- ++(0.4,0) |- (srt.west);
\draw[arrow, gray!70] (j2.east) -- (srt.west);
\draw[arrow, gray!70] (j3.east) -- ++(0.4,0) |- (srt.west);

\begin{scope}[on background layer]
  \node[draw=black, fill=none, dashed, thick,
        rounded corners=6pt,
        fit=(vramprobe)(poolbuilder)(queue)(f1)(f2)(f3)(fdots)
            (t1)(t2)(t3)(j1)(j2)(j3)(srt),
        inner sep=12pt, label={[audiogreen!70, font=\small\bfseries]
        north:Single-GPU path (ThreadPoolExecutor)}] {};
\end{scope}

\end{tikzpicture}
\caption{VRAM-aware transcription pipeline. A probe measures the free GPU memory, a builder fills a pool with as many Whisper instances as fit under a configurable headroom, and a thread pool of workers consumes the pool to transcribe MP3 chunks in parallel. Each completed file is merged into a single SRT file with absolute UTC timestamps.}
\label{fig:transcription_parallel}
\end{figure}

Each MP3 chunk is transcribed by Whisper large-v3~\cite{radford2023whisper}, a large-scale ASR foundation model pre-trained on 680{,}000 hours of multi-domain speech. Whisper and Wav2Vec~2.0 are the two voice encoders shown in the voice tower of Fig.~\ref{fig:arch_variations}. The same encoder is reused for two purposes: (i) producing the text transcription used by the callsign-matching logic of Section~\ref{sec:linking} and (ii) producing the voice embedding $\vemb$ consumed by the contrastive stage. Using a single high-capacity encoder for both roles keeps the pipeline simple and ensures that the embedding used downstream is exactly the one from which the transcription is decoded, so a point in the voice latent space can in principle be turned back into a plausible utterance through the same network. The choice of Whisper over Wav2Vec~2.0~\cite{baevski2020wav2vec2}, namely a community-fine-tuned ATC-specific variant, is justified in Appendix~C on the basis of (i) the qualitative structure of the latent space, (ii) the quality of the transcriptions obtained when decoding from the embedding, and (iii) a sweep over projector sizes.

\emph{Voice Activity Detection (VAD)} is run before transcription using Silero VAD, with parameters tuned for the short, clipped style of ATC speech: detection threshold $0.3$ (lower than the default $0.5$), minimum speech duration \SI{50}{\milli\second}, and \SI{800}{\milli\second} of pad on either side of each detected segment. These settings recover short clearances that the default settings would discard, while keeping false positives manageable.

In practice, transcription was by far the most time-consuming stage of the entire study and the main bottleneck of the pipeline. The difficulty is twofold. Firstly, Whisper large-v3 is slow to run over the full volume of archived audio, and secondly its GPU memory footprint is incompatible with naive parallel processing, so simply launching many instances at once quickly exhausts the available VRAM. We thus adopt a VRAM-aware model pool, sketched in Fig.~\ref{fig:transcription_parallel}. Whisper instances are loaded one by one and the process stops as soon as the remaining free memory drops below a configurable headroom (typically \SI{4}{\giga\byte}). The resulting pool of $N$ instances is consumed by a thread pool, with each worker grabbing a model, transcribing a single MP3, and returning the model to the pool. Outputs are written atomically through a \texttt{.tmp} rename so that a partial run can be resumed without re-transcribing files already on disk. Inference uses half-precision float on GPU and integer quantization on CPU as a fallback.

A transcribed line of ATC speech such as ``United two three four, descend and maintain five thousand'' carries the callsign of the addressed aircraft. We extract this information through a cascade of regular-expression patterns that go from most to least specific:

\begin{center}
\small
\begin{tabular}{lll}
\toprule
Pattern type            & Example match & Priority \\
\midrule
Full callsign           & \texttt{UAL 234}   & 1000 \\
Telephony + suffix      & ``United two three four'' & 950 \\
Suffix only             & ``two three four'' & 900 \\
Suffix partial          & ``two three ''     & 100 \\
Telephony designator    & ``United''         & 50 \\
\bottomrule
\end{tabular}
\end{center}

\begin{figure}[htbp]
\centering
\begin{subfigure}{0.48\textwidth}
  \includegraphics[width=\textwidth]{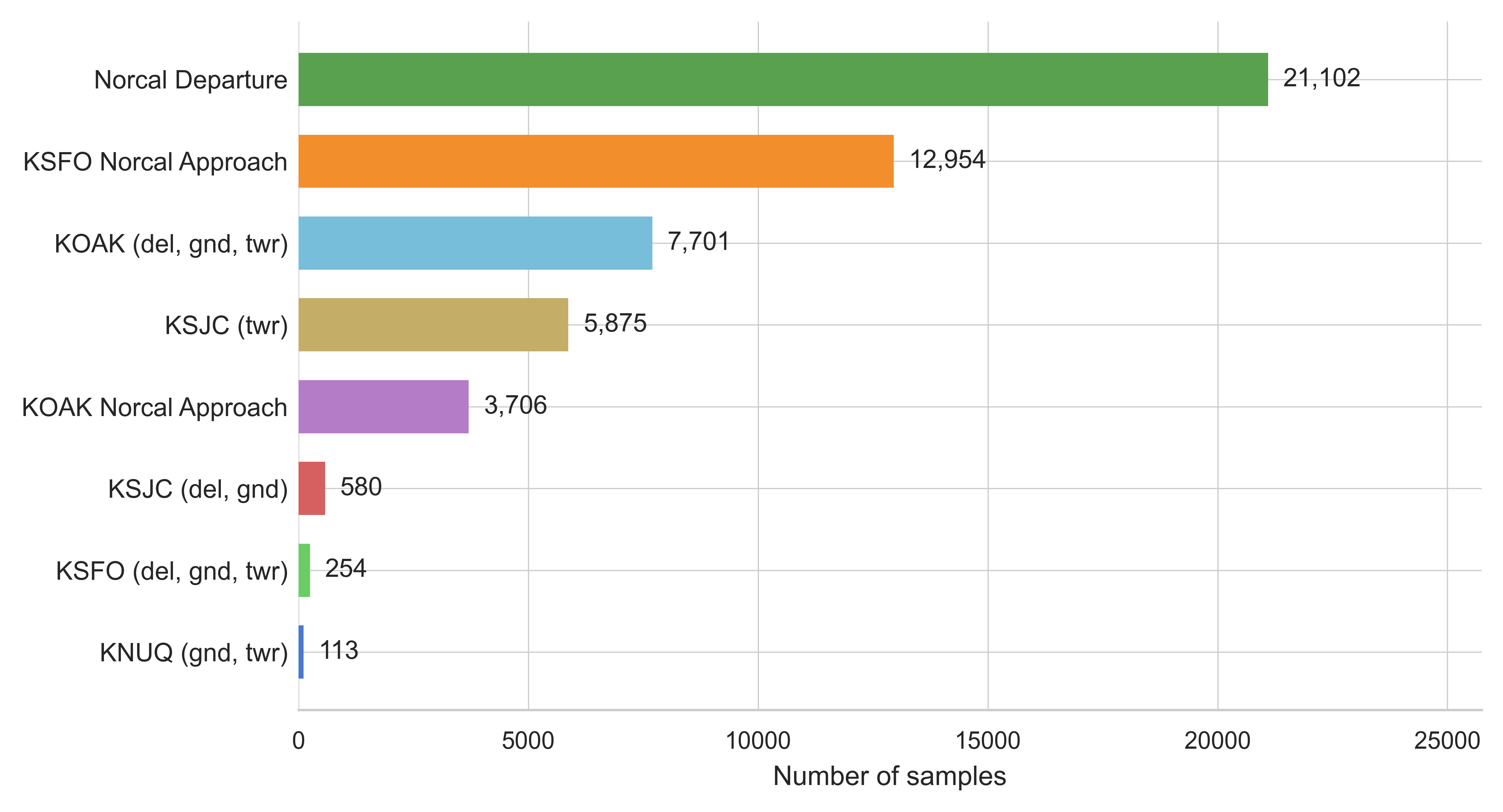}
  \caption{Sample counts per ATC frequency.}
\end{subfigure}\hfill
\begin{subfigure}{0.48\textwidth}
  \includegraphics[width=\textwidth]{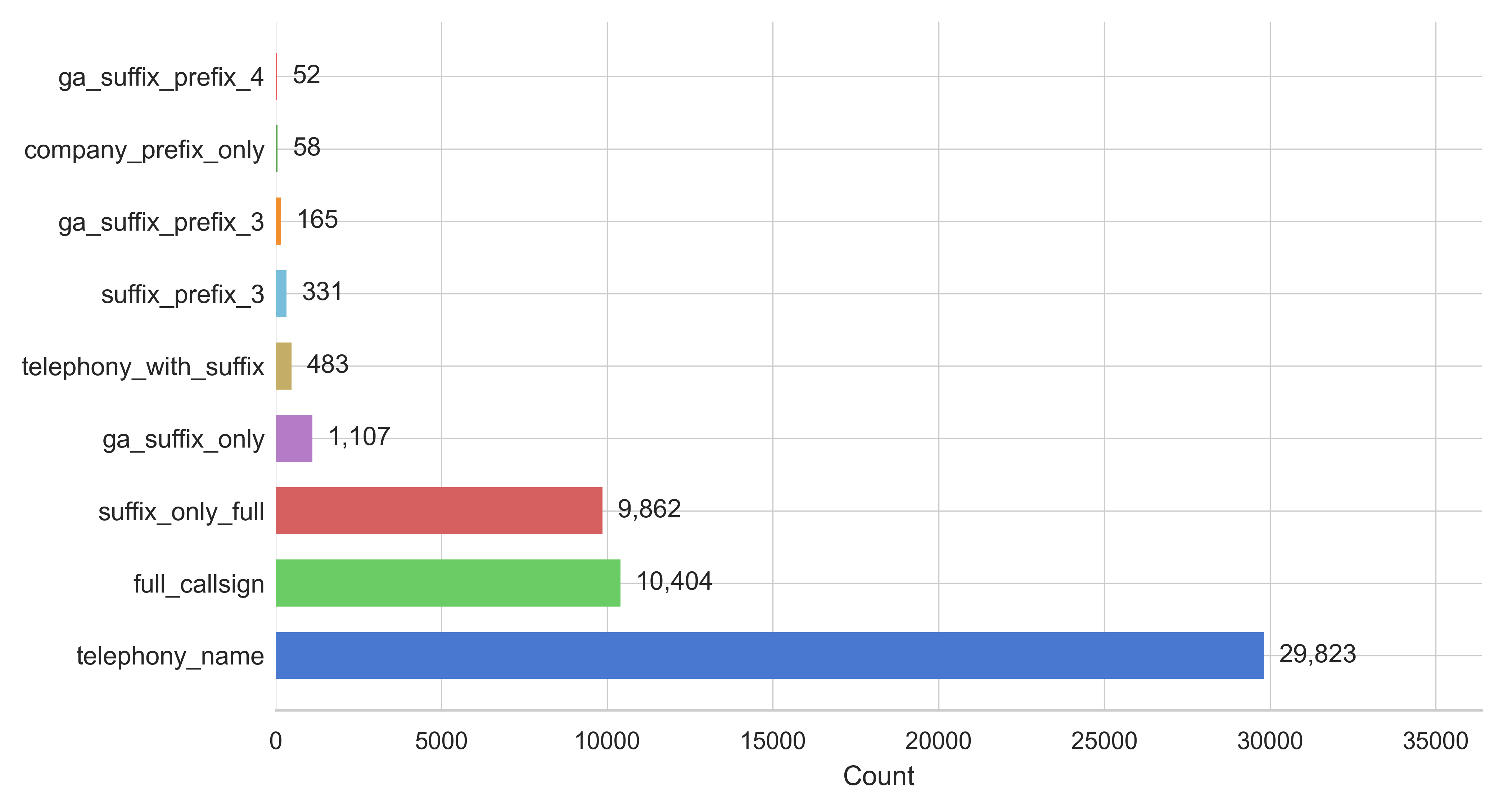}
  \caption{Distribution of callsign matching strategies.}
\end{subfigure}
\caption{Composition of the paired voice--trajectory dataset.}
\label{fig:match_dataset}
\end{figure}

Patterns tolerate arbitrary whitespace and hyphens between characters, because ASR systems frequently insert separators inside identifiers (e.g.\ ``U-A-L 2-3-4''). Lookups across consecutive transcription lines are also performed, since a callsign can be split across two transmissions. Telephony designators are normalized through a curated mapping, directly scraped from this \href{https://www.faa.gov/air_traffic/publications/atpubs/cnt_html/chap3_section_3.html}{FAA website} (e.g.\ ``Air France'' $\to$ \texttt{AFR}), and digit words are converted from English to numerals (e.g.\ ``two three four'' $\to$ \texttt{234}). When several patterns match the same line, only the highest-priority match is kept. The distribution of match types in the final dataset is shown in Fig.~\ref{fig:match_dataset}.

\subsection{Linking Modalities}
\label{sec:linking}

\begin{figure}[bp]
\centering
\begin{tikzpicture}[
  node distance = 0.6cm and 1.1cm,
  box/.style = {
    rounded corners=4pt, draw, thick,
    minimum width=2.6cm, minimum height=1.0cm,
    align=center, font=\small\bfseries,
    text=white
  },
  arrow/.style = {-{Stealth[length=6pt]}, thick, gray!70},
  label/.style = {font=\footnotesize\itshape, text=darkgray, align=center},
  rowlabel/.style = {font=\footnotesize\bfseries, text=black, align=right},
]

\node[box, fill=adsbblue]                       (adsb)   {ADS-B\\Filtering};
\node[box, fill=audiogreen, right=of adsb]      (audio)  {Audio\\Transcription};
\node[box, fill=linkpurple, right=of audio]     (link)   {Multimodal\\Linking};
\node[box, fill=jointorange, right=of link]     (joint)  {Joint\\Assembly};

\draw[arrow] (adsb)  -- (audio);
\draw[arrow] (audio) -- (link);
\draw[arrow] (link)  -- (joint);

\node[label, above=0.35cm of adsb]  (in_adsb)  {ADS-B CSV files};
\node[label, above=0.35cm of audio] (in_audio) {ATC MP3 files};

\node[label, below=0.35cm of adsb]  (out_adsb)  {NetCDF4\\trajectories};
\node[label, below=0.35cm of audio] (out_audio) {SRT files\\(UTC timestamps)};
\node[label, below=0.35cm of link]  (out_link)  {Callsign\\matches JSON};
\node[label, below=0.35cm of joint] (out_joint) {HDF5 shards\\(waveform + traj)};

\node[rowlabel, left=0.4cm of in_adsb.west] {Input data};
\node[rowlabel, left=0.4cm of adsb.west]    {Processing};
\node[rowlabel, left=0.4cm of out_adsb.west]{Output files};

\end{tikzpicture}
\caption{Linking pipeline from raw sources to joint HDF5 shards. The first row lists the inputs, the second row the processing stages, and the third row the intermediate or final output files.}
\label{fig:pipeline_overview}
\end{figure}
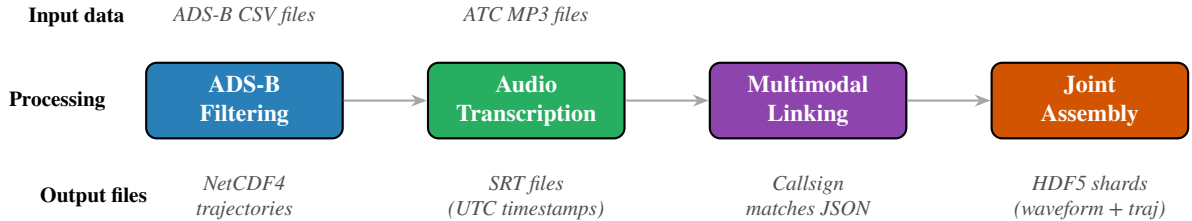

The overall linking pipeline takes the raw ADS-B and audio streams and produces the final HDF5 shards. Its inputs, intermediate processes, and outputs are summarized in Fig.~\ref{fig:pipeline_overview}. The pairing itself operates flight-first. For every ADS-B segment identified by its ICAO~24 code, callsign, and time window $[t_\text{start}, t_\text{end}]$, we search the transcription file for all lines whose callsign matches and whose timestamp falls within $[t_\text{start} - \Delta_\text{pre}, t_\text{end} + \Delta_\text{post}]$ with $\Delta_\text{pre} = \Delta_\text{post} = \SI{60}{\second}$. This direction is significantly more robust than the opposite one (searching for flights given a transcribed callsign) because the temporal constraint eliminates false positives from other flights with similar callsigns operating elsewhere.

\begin{figure}[htbp]
\centering
\begin{subfigure}{0.85\textwidth}
  \includegraphics[width=\textwidth]{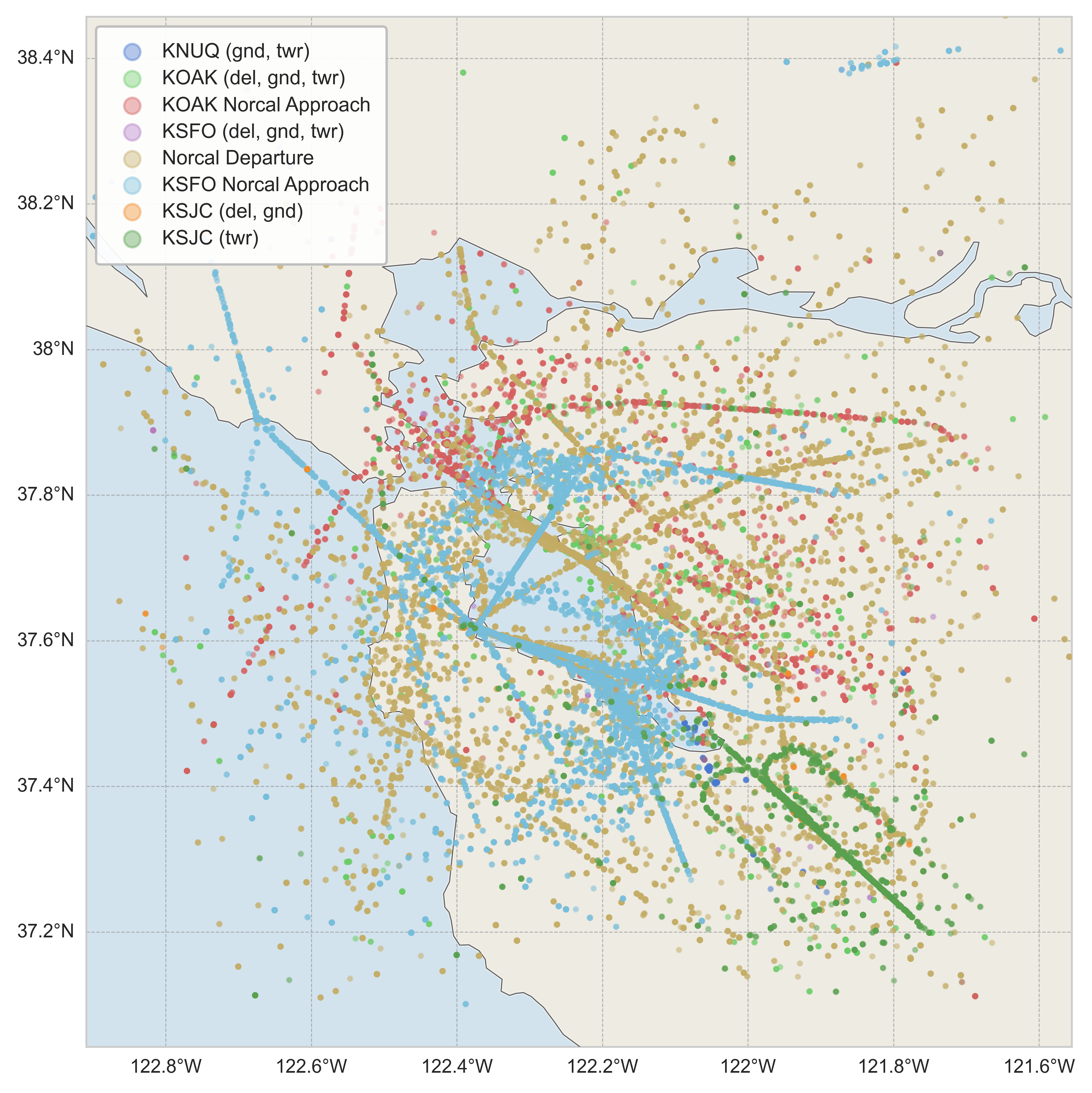}
  \caption{Aircraft position at sentence start, colored by ATC frequency. One dot per flight over the four-day collection window.}
\end{subfigure}
\\[0.8em]
\begin{subfigure}{0.95\textwidth}
  \includegraphics[width=\textwidth]{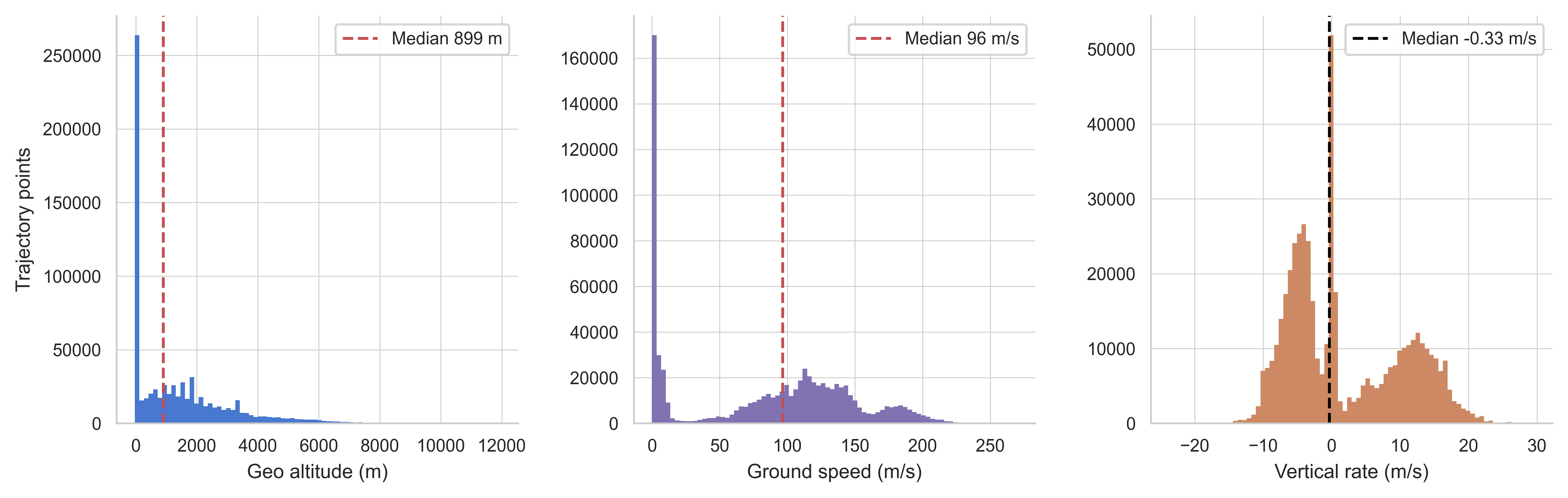}
  \caption{Distributions of altitude, ground speed, and vertical rate over all trajectory points. The joint dataset contains a large number of data points with zero altitude, zero speed, and zero vertical rate, since most communications take place on the ground during taxiing.}
\end{subfigure}
\caption{Statistical overview of the paired joint dataset over the San Francisco Bay Area.}
\label{fig:dataset_overview}
\end{figure}

Each matched pair produces a sample of the joint dataset. The sample contains both modalities and their metadata, namely (i) a raw audio clip extracted from the MP3 at the matched timestamps, with a \SI{750}{\milli\second} tail pad to capture the end of each transmission cleanly; (ii) a 14-step trajectory window centered on the speech event (nine points before, current, four after); and (iii) a metadata record with the callsign, the matched timestamps, the match type, the source frequency, and the source MP3 filename. The data is serialized into chunked HDF5 shards organized by tower, with audio stored as variable-length \texttt{int16} arrays at \SI{16}{\kilo\hertz} and trajectories stored as \texttt{float32} arrays of shape $(N, T, 7)$ where the seventh column carries the absolute timestamp. The high-level statistics of the resulting joint corpus are reported in Table~\ref{tab:dataset_summary}, the per-frequency sample distribution in Table~\ref{tab:tower_counts}, and the geographic footprint together with the marginal distributions of altitude, ground speed, and vertical rate in Fig.~\ref{fig:dataset_overview}.

\begin{table}[hbt!]
\caption{\label{tab:dataset_summary}Summary of the paired voice--trajectory dataset over the San Francisco Bay Area.}
\centering
\begin{tabular}{lr}
\toprule
Metric & Value \\
\midrule
Control tower frequencies (KSFO / KOAK / KSJC / Moffett) & 8 \\
Unique aircraft (ICAO24 codes) & 962 \\
Unique callsigns & 1{,}411 \\
Paired voice--trajectory samples & 52{,}285 \\
Trajectory window length $T$ (timesteps) & 14 \\
Median ADS-B sampling interval $\Delta t$ (s) & 27 \\
Median phrase duration (s) & 6.0 \\
Median transcription length (words) & 12 \\
Median altitude (m) & 899 \\
Median ground speed (m\,s$^{-1}$) & 96 \\
\bottomrule
\end{tabular}
\end{table}
\begin{table}[hbt!]
\caption{\label{tab:tower_counts}Distribution of paired voice--trajectory samples across the eight monitored ATC frequencies. The three NorCal en-route frequencies together account for more than 72\,\% of the dataset.}
\centering
\begin{tabular}{lrr}
\toprule
ATC frequency / position & Samples & \% of total \\
\midrule
\texttt{ksfo\_koak\_norcal\_dep} (NorCal Departure)         & 21{,}102 & 40.4 \\
\texttt{ksfo\_norcal\_app2\_l} (NorCal Approach L)          & 12{,}954 & 24.8 \\
\texttt{koak\_del\_gnd\_twr} (KOAK Delivery / Ground / Twr) &  7{,}701 & 14.7 \\
\texttt{ksjc\_twr2} (KSJC Tower)                            &  5{,}875 & 11.2 \\
\texttt{koak\_norcal\_app} (KOAK NorCal Approach)           &  3{,}706 &  7.1 \\
\texttt{ksjc\_del\_gnd} (KSJC Delivery / Ground)            &    580   &  1.1 \\
\texttt{ksfo\_del\_gnd\_alt} (KSFO Delivery / Ground alt.)  &    254   &  0.5 \\
\texttt{knuq\_gnd\_twr} (KNUQ Ground / Tower)               &    113   &  0.2 \\
\midrule
\textbf{Total}                                              & \textbf{52{,}285} & 100 \\
\bottomrule
\end{tabular}
\end{table}

\begin{figure}[htbp]
\centering
\includegraphics[width=0.9\textwidth]{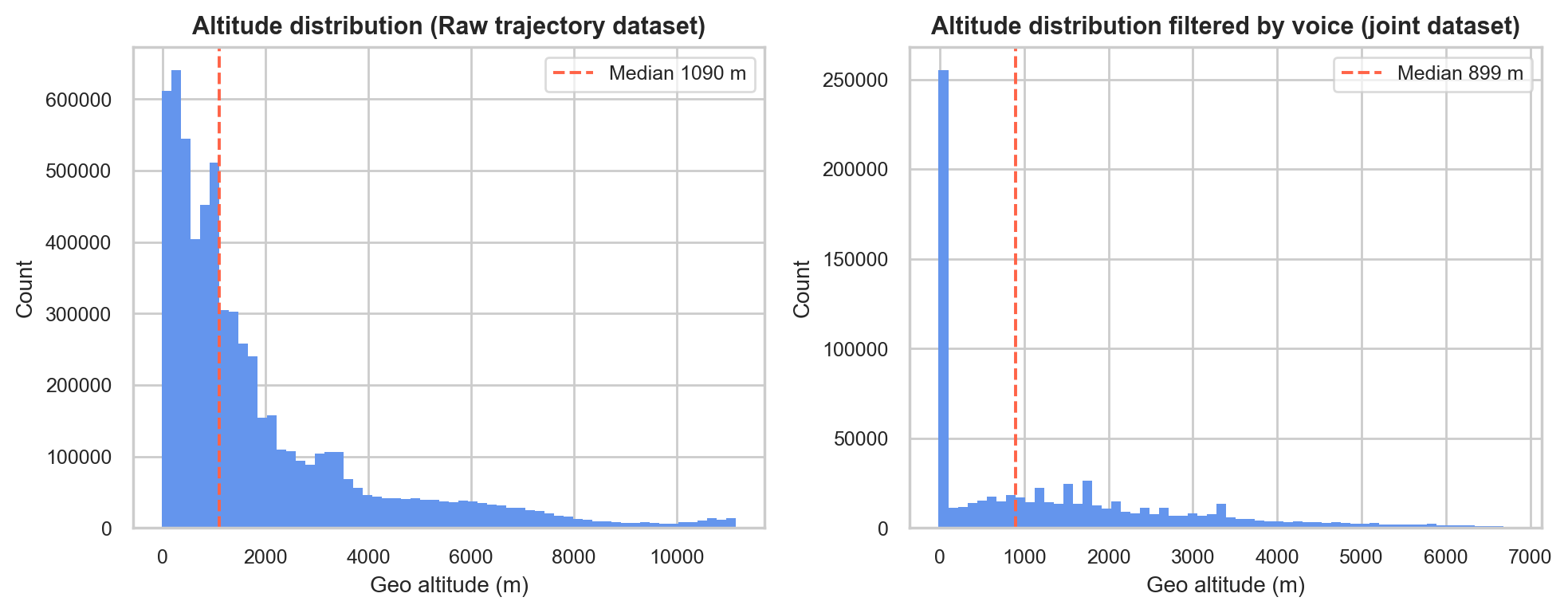}
\caption{Evolution of the altitude distribution as the dataset moves from the raw ADS-B feed to the cleaned trajectory dataset and finally to the joint voice--trajectory dataset. Each pairing stage discards data points that cannot be matched, which shrinks the effective dataset and biases it toward instrumented commercial traffic.}
\label{fig:altitude_shrinkage}
\end{figure}

A side effect of the linking stage is a drastic reduction of the effective dataset size. The raw ADS-B feed contains millions of trajectory points, but only the segments that can be matched to a transcribed callsign within the temporal window survive into the joint dataset. Figure~\ref{fig:altitude_shrinkage} illustrates how the altitude distribution evolves as the data move along the pipeline, from the raw ADS-B data to the cleaned trajectory dataset and finally to the paired voice--trajectory dataset. The general shape of the distribution is preserved, but the dataset becomes increasingly biased toward operations that produce transcribable voice activity, namely commercial flights on instrumented sectors. This bias is a known limitation of the current corpus and motivates the multi-source extension discussed in Section~\ref{sec:conclusion}.

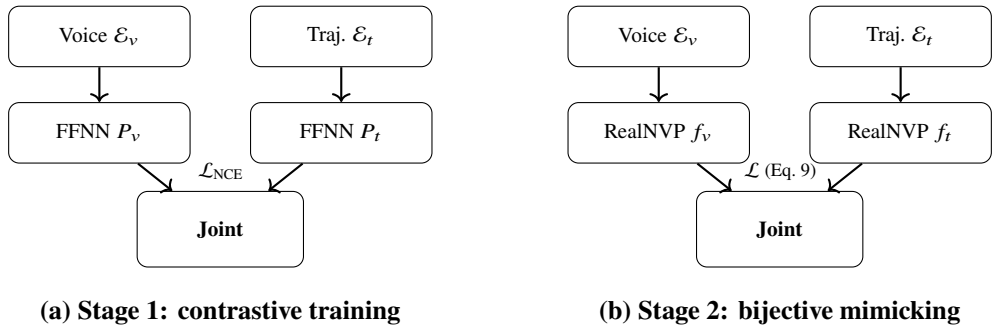
\begin{figure}[bp]
\centering
\begin{tikzpicture}[
  yscale=0.85,
  enc/.style={draw, rounded corners, minimum width=2.4cm, minimum height=0.8cm, align=center, font=\footnotesize},
  proj/.style={draw, rounded corners, minimum width=2.4cm, minimum height=0.8cm, align=center, font=\footnotesize},
  flow/.style={draw, rounded corners, minimum width=2.4cm, minimum height=0.8cm, align=center, font=\footnotesize},
  joint/.style={draw, rounded corners, minimum width=2.2cm, minimum height=1.0cm, align=center, font=\footnotesize\bfseries},
  arr/.style={->, thick},
  lbl/.style={font=\scriptsize, inner sep=2pt},
]
\begin{scope}[shift={(0,0)}]
\node[enc]   (v1)  at (-1.6, 3.0) {Voice $\Ev$};
\node[enc]   (t1)  at ( 1.6, 3.0) {Traj.\ $\Et$};
\node[proj]  (pv1) at (-1.6, 1.5) {FFNN $\Pv$};
\node[proj]  (pt1) at ( 1.6, 1.5) {FFNN $\Pt$};
\node[joint] (j1)  at (0, 0) {Joint};
\draw[arr] (v1)  -- (pv1);
\draw[arr] (t1)  -- (pt1);
\draw[arr] (pv1) -- (j1);
\draw[arr] (pt1) -- (j1);
\node[lbl] at (0, 0.9) {$\Lloss_{\text{NCE}}$};
\node at (0, -1.3) {\textbf{(a) Stage 1: contrastive training}};
\end{scope}
\begin{scope}[shift={(7.4,0)}]
\node[enc]   (v2) at (-1.6, 3.0) {Voice $\Ev$};
\node[enc]   (t2) at ( 1.6, 3.0) {Traj.\ $\Et$};
\node[flow]  (fv) at (-1.6, 1.5) {RealNVP $\fv$};
\node[flow]  (ft) at ( 1.6, 1.5) {RealNVP $\ft$};
\node[joint] (j2) at (0, 0) {Joint};
\draw[arr] (v2) -- (fv);
\draw[arr] (t2) -- (ft);
\draw[arr] (fv) -- (j2);
\draw[arr] (ft) -- (j2);
\node[lbl] at (0, 0.9) {$\Lloss$ (Eq.~\ref{eq:nvp_loss})};
\node at (0, -1.3) {\textbf{(b) Stage 2: bijective mimicking}};
\end{scope}
\end{tikzpicture}
\caption{Two-stage training pipeline. In stage~(a) the two FFNN projectors are trained jointly by the InfoNCE objective. The frozen joint-space targets are then used in stage~(b) to train two normalizing flows that mimic a bijective map between each modality space and the joint space. For stage~(b), synthetic data generated by direct inference are used in addition to real data to increase the representativeness of the sampling of the initial space.}
\label{fig:pipeline}
\end{figure}

The joint architecture is trained in two stages, as illustrated in Fig.~\ref{fig:pipeline}.

\subsubsection*{Stage 1: Contrastive Training} The two FFNN projectors are trained jointly with the symmetric InfoNCE objective of Eq.~\eqref{eq:infonce}. The upstream encoders $\Ev, \Et$ are frozen. Each projector has a single hidden layer of width 4096 with a SELU activation and a dropout of $0.2$, followed by a linear output layer. We use the AdamW optimizer with learning rate $10^{-4}$, weight decay $10^{-4}$, batch size $B = 512$, and gradient clipping at norm $1.0$. The temperature $\tau$ is parameterized as $\tau = \exp(\theta)$ with $\theta$ learnable and clamped to $\tau \in [0.04, 100]$ to prevent collapsing or explosion. Learning-rate scheduling is performed by \texttt{ReduceLROnPlateau} on the validation loss, with patience seven and factor $0.5$, and training stops when no improvement is seen for fifteen consecutive epochs. The complete set of hyperparameters is summarized in Table~\ref{tab:contrastive_hparams}.

\subsubsection*{Stage 2: Bijective Lifting} The contrastive projectors are frozen and used as ground-truth oracles. The two normalizing flows are then trained with the loss in Eq.~\eqref{eq:nvp_loss}, with weights $\lambda_f = \lambda_b = 0.5$ and $\lambda_c = 0.1$. Each flow contains eight affine coupling blocks with conditioner MLPs of hidden width $[256, 256]$, interleaved with ActNorm layers. Because the real paired data only cover a limited region of the embedding space, the training set is augmented with $4 \times 10^4$ synthetic embeddings per modality: half drawn from the per-dimension bounding box of real data, half from Gaussian perturbations centered on random real samples with $\sigma_d = 0.5 \times \text{std}_d$. Synthetic embeddings receive their joint-space target by passing through the frozen contrastive projector. The optimizer is again AdamW with learning rate $10^{-4}$, weight decay $10^{-4}$, and batch size 256.

\begin{table}[hbt!]
\caption{\label{tab:contrastive_hparams}Hyperparameters of the contrastive training stage (Stage~1).}
\centering
\begin{tabular}{ll}
\toprule
Component & Configuration \\
\midrule
Voice projector $\Pv$       & FFNN $1280 \to 4096 \to 1024$, SELU, dropout 0.2 \\
Trajectory projector $\Pt$  & FFNN $1792 \to 4096 \to 1024$, SELU, dropout 0.2 \\
Optimizer                   & AdamW, $\beta_1 = 0.9$, $\beta_2 = 0.999$ \\
Initial learning rate       & $10^{-4}$ \\
Weight decay                & $10^{-4}$ \\
Gradient clipping (max norm) & 1.0 \\
LR scheduler                & \texttt{ReduceLROnPlateau}, factor 0.5, patience 7 \\
Early stopping patience     & 15 epochs ($\delta_{\min} = 10^{-4}$) \\
Batch size                  & 512 \\
Temperature bounds          & $\tau \in [0.04, 100]$ (learned in log-space) \\
Max epochs                  & 300 \\
\bottomrule
\end{tabular}
\end{table}

\textit{Network capacity vs.\ data scale}

The two projectors contain on the order of $10^7$ trainable parameters each: the voice projector has approximately $1280 \cdot 4096 + 4096 \cdot 1024 \approx 9.5 \times 10^6$ weights, and the trajectory projector approximately $1792 \cdot 4096 + 4096 \cdot 1024 \approx 1.1 \times 10^7$ weights. The two normalizing flows together carry a comparable number of parameters, dominated by the conditioner MLPs of the coupling layers. The validation set used for early stopping contains on the order of $5 \times 10^3$ paired samples. A sample here is not a scalar observation. On the trajectory side alone it carries a $14 \times 7$ window of seven recorded channels, and it is paired with a full voice clip, so the alignment has to satisfy on the order of $10^5$ scalar constraints rather than $5 \times 10^3$. That figure remains below the parameter count, and the projectors are therefore large relative to the data they see. What keeps the optimization well behaved is not the ratio itself but the regularization around it, namely the dropout, the weight decay, and the early stopping that stop the projectors from memorizing the validation set. The persistent train--validation gap observed in Section~\ref{sec:results_retrieval} indicates that a further reduction of projector width would be a sensible follow-up, which we did not pursue here because retrieval quality is already informative enough to support the downstream studies.

\section{Results}
\label{sec:results}
We evaluate V2TATC on six tasks that probe different aspects of the joint space.

\subsection{Cross-Modal Retrieval}
\label{sec:results_retrieval}
\begin{figure}[bp]
\centering
\begin{subfigure}{0.48\textwidth}
  \includegraphics[width=\textwidth]{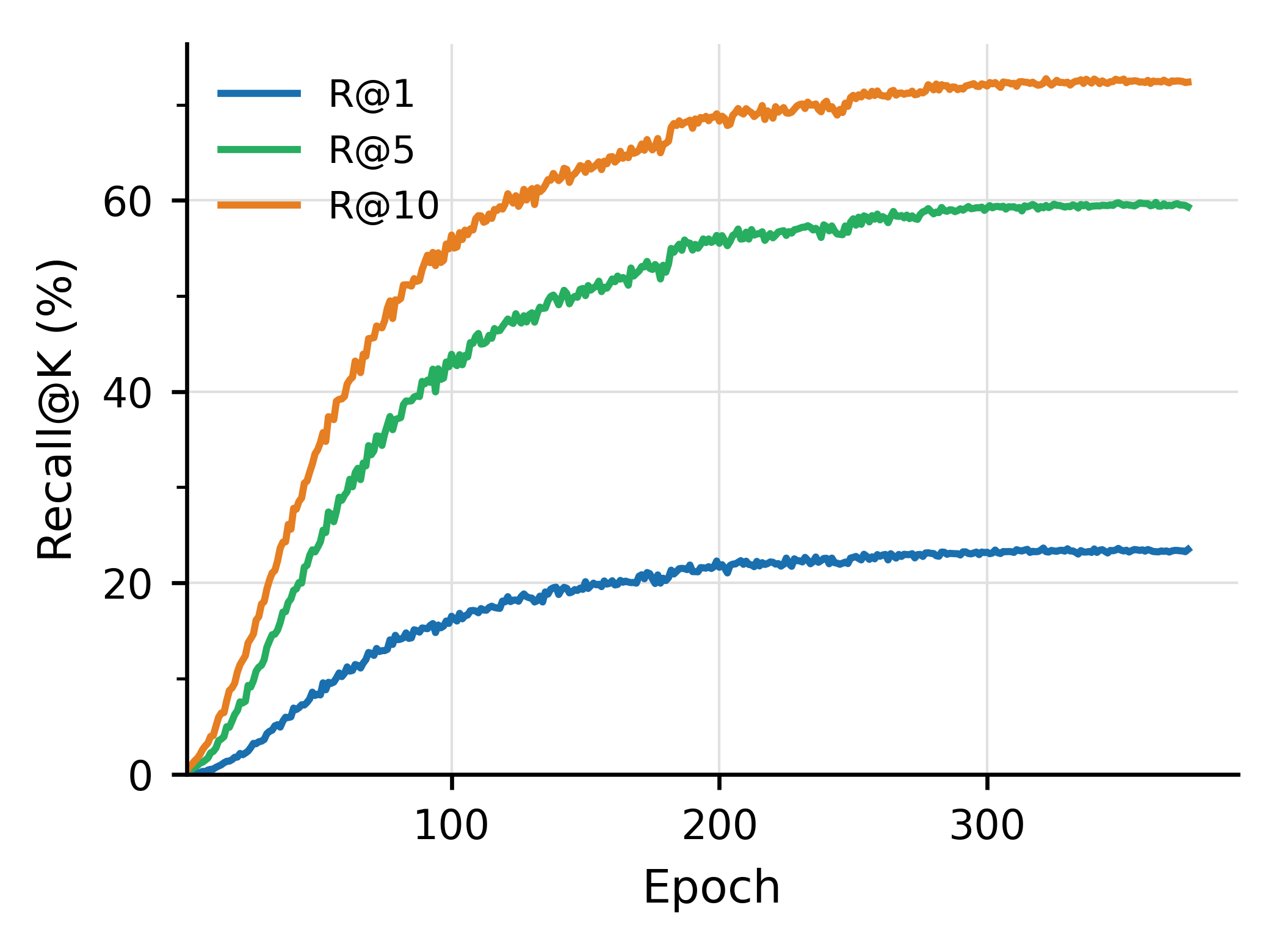}
  \caption{Recall@$K$ on the $5{,}229$ validation candidates over training epochs. R@1, R@5 and R@10 increase steadily and saturate after about 200 epochs at final values $23.5\%$, $59.7\%$ and $72.6\%$, three orders of magnitude above the random retrieval baseline ($\text{R@1} \approx 0.02\%$).}
\end{subfigure}\hfill
\begin{subfigure}{0.48\textwidth}
  \includegraphics[width=\textwidth]{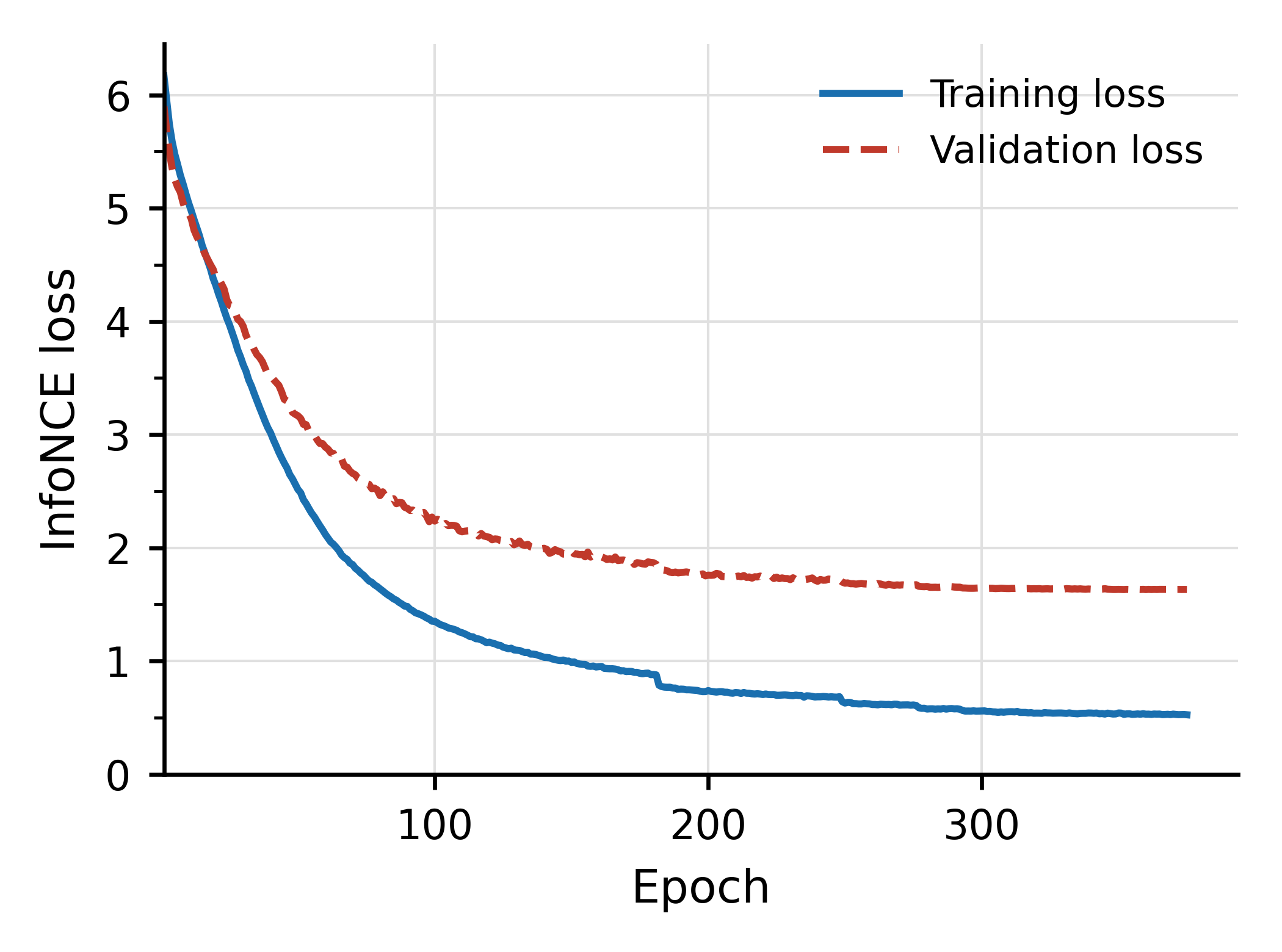}
  \caption{Symmetric InfoNCE loss during training. The training loss (solid blue) reaches a minimum of approximately $0.45$, while the validation loss (dashed red) stabilizes near $1.59$, reflecting the limited diversity of the paired dataset rather than overfitting. Both curves are averaged over the voice-to-trajectory and trajectory-to-voice directions.}
\end{subfigure}
\caption{Contrastive training dynamics: retrieval performance (a) and InfoNCE loss (b) over training epochs.}
\label{fig:retrieval}
\end{figure}

We first measure cross-modal retrieval. Given a voice query in the validation set, we rank all candidate trajectories by cosine similarity in the joint space and record whether the true trajectory falls in the top $K$. The metric \emph{Recall at $K$} (R@$K$) is the fraction of queries for which this is the case. R@$K$ values are reported in Table~\ref{tab:retrieval} and the corresponding training curves are shown in Fig.~\ref{fig:retrieval}.

\begin{table}[hbt!]
\caption{\label{tab:retrieval}Cross-modal retrieval performance in the joint space (voice query, trajectory candidate, $N = 5{,}229$ validation candidates). Recall values are percentages; InfoNCE loss and temperature are dimensionless. The trained model improves over random retrieval by roughly three orders of magnitude on R@1.}
\centering
\begin{tabular}{lccccc}
\toprule
Metric                                          & R@1 (\%)  & R@5 (\%)  & R@10 (\%)  & InfoNCE validation loss  & Learned temperature $\tau$  \\
\midrule
Random baseline                                 & 0.02      & 0.10      & 0.19       & $\log(B=512)$ = 6.24                    & -- \\
Ours (validation)              & 23.5      & 59.7      & 72.6       &  1.585         & 0.04 (floor of clamp) \\
\midrule
Improvement factor                              & $\times 1175$ & $\times 597$ & $\times 382$ & $\div 4$ &  \\
\bottomrule
\end{tabular}
\end{table}

The trained model achieves R@1 of $0.235$ and R@10 of $0.726$ over $5{,}229$ validation candidates. The random baseline is $K / N_\text{val} \approx 2 \times 10^{-4}$ for $K=1$. The model therefore outperforms random retrieval by roughly three orders of magnitude, as quantified explicitly in Table~\ref{tab:retrieval}. The fact that R@10 is roughly three times R@1 suggests that the model frequently confuses the correct trajectory with a few similar candidates (typically aircraft of the same operator on the same sector), but it almost always recovers the right region of the joint space. Thus, this analysis illustrates that our bijection is indeed established between submanifolds of our two spaces and not directly between pairs of samples.

\begin{figure}[tp]
\centering
\begin{subfigure}{0.38\textwidth}
  \includegraphics[width=\textwidth]{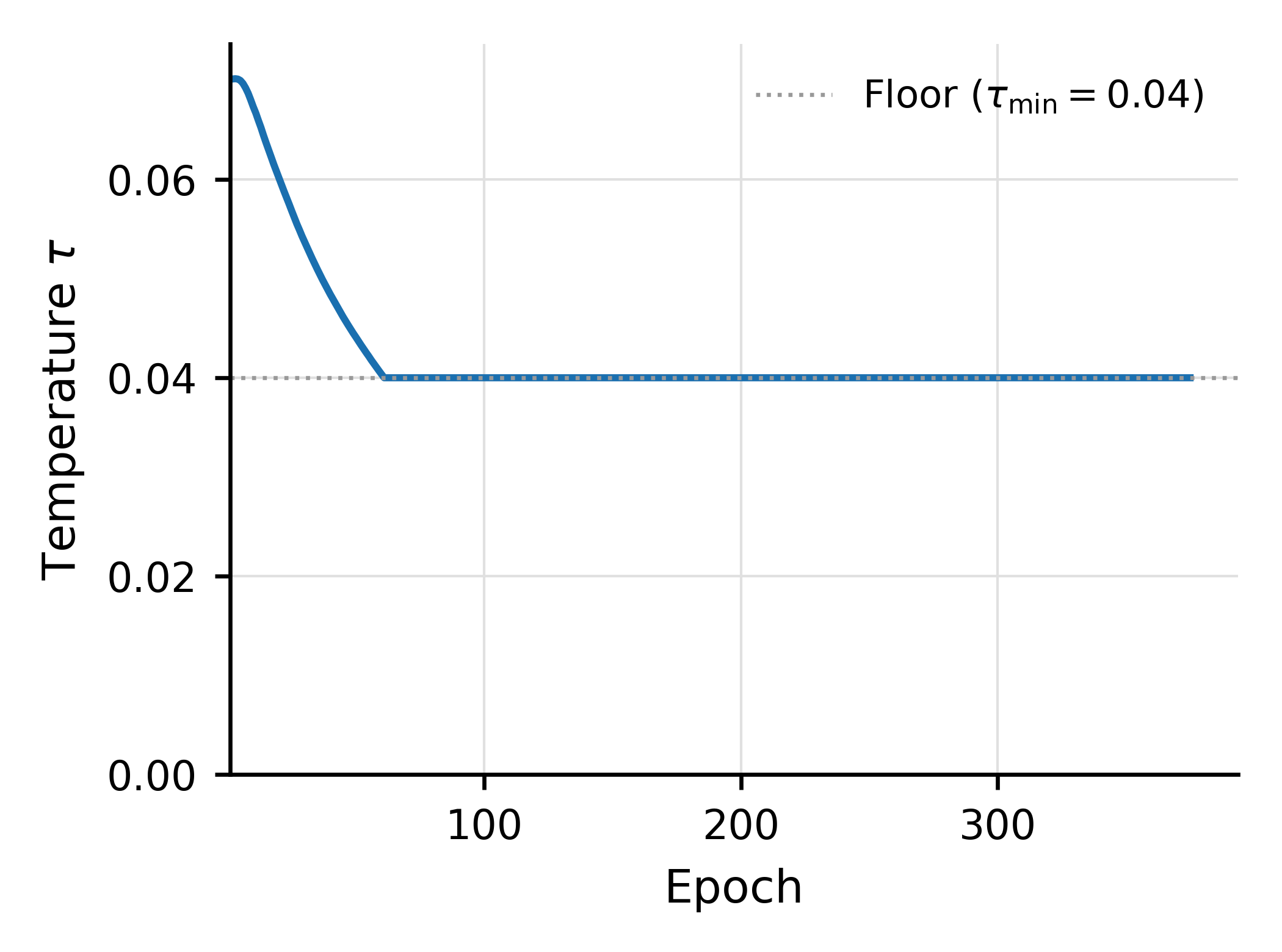}
  \caption{Evolution of the learnable contrastive temperature $\tau$ (parameterized in log-space). The temperature drops abruptly to its lower bound $\tau_{\min}=0.04$ (dotted line) within the first few epochs and remains pinned there. The abrupt collapse reads as a signature of the limited diversity of the data, which provides a small set of clearly separable directions sufficient to drive the loss sharply downward.}
\end{subfigure}\hfill
\begin{subfigure}{0.58\textwidth}
  \includegraphics[width=\textwidth]{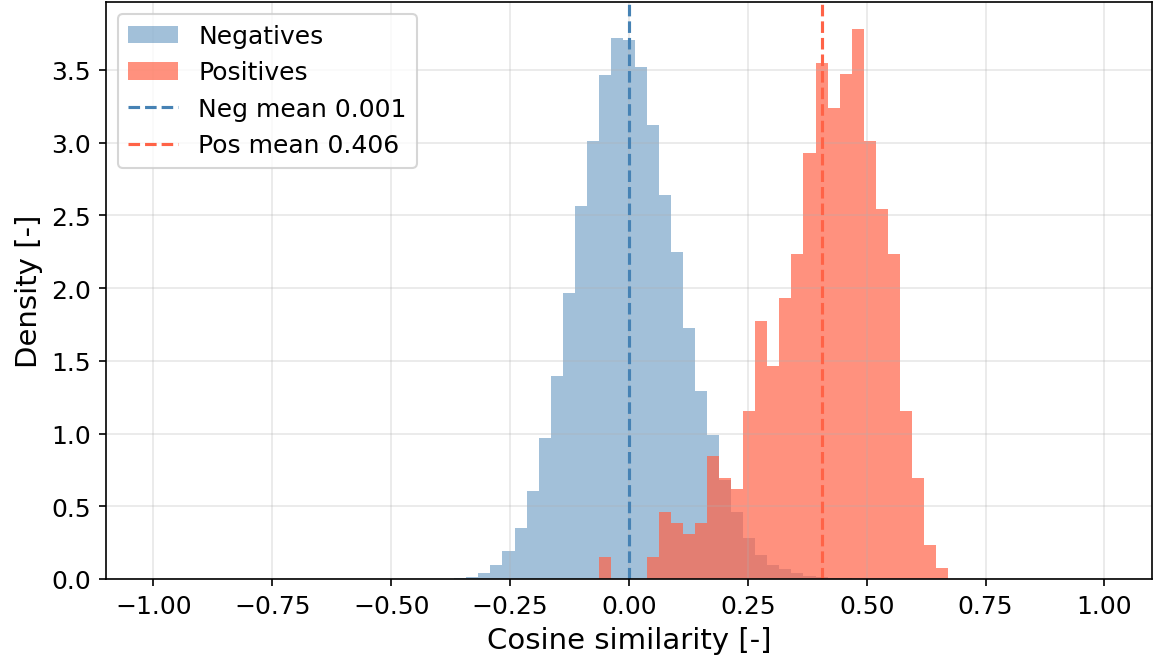}
  \caption{Distribution of cosine similarities between voice and trajectory embeddings projected into the joint space. Positive (matched) pairs should tend toward~$1$ and negative (mismatched) pairs concentrate near~$0$. The small overlap corresponds mainly to aircraft of the same operator on the same sector, which produce structurally similar voice phrases and motion profiles.}
\end{subfigure}
\caption{Sharpness of the contrastive alignment: temperature evolution (a) and cosine similarity distributions (b).}
\label{fig:temperature}
\end{figure}

The learned temperature $\tau$ drifts down to its floor value of $0.04$ during training (Fig.~\ref{fig:temperature}), which is consistent with the InfoNCE literature, where a small temperature sharpens the softmax distribution and pushes harder on negatives. The clamp interval $[0.04, 100]$ used here matches the CLIP defaults~\cite{radford2021clip}, and the steady drift toward the lower bound is the expected behavior on a well-aligned dataset. However, the way $\tau$ hits the floor in our setting is informative on its own. Instead of decreasing smoothly along the validation loss, it drops abruptly within a few epochs and then stays pinned at the floor. Recent analyses of InfoNCE~\cite{infonce_gap_2024} show that the assumptions behind its theoretical guarantees rarely hold in practice, and in particular that the latent factors of positive pairs vary to very unequal extents. We read the abrupt drop as a signature of the dataset rather than of the architecture. The dataset does not sample the embedding space uniformly enough, so a small number of clearly separable directions are enough to drive the loss down sharply, and the model has no incentive to keep $\tau$ above the clamp. The persistent train/validation gap (training loss $\approx 0.45$ vs.\ validation loss $\approx 1.59$ at the best epoch) is consistent with this reading, and the limiting factor at this stage is the diversity of the paired dataset rather than the capacity of the projectors.

\subsection{Trajectory Forecast}
\label{sec:results_forecast}
We next evaluate the accuracy of trajectory forecasting. The task is to predict the next $H = 4$ ADS-B steps given $T - H = 10$ steps of context. We compare two paths. Path~A uses the MAE encoder directly, with the last four timesteps masked and reconstructed by the reconstruction head trained during MAE pre-training. Path~B uses the cross-modal chain voice $\rightarrow$ joint $\rightarrow$ trajectory. A voice phrase is mapped to the trajectory side through $\fv$ then $\ft^{-1}$, and a small learned head (the StepPredictor) predicts the next four steps from the recovered representation. Path~A is therefore a pure surveillance-based forecast and serves as an upper bound. Path~B measures how much of that capability is preserved when the only input is a voice phrase.

\begin{table}[hbt!]
\caption{\label{tab:forecast}Short-horizon trajectory forecast (next four ADS-B steps from ten-step context). MSE values are in normalized delta z-score space (dimensionless).}
\centering
\begin{tabular}{lcccc}
\toprule
Path & Global MSE & Position MSE & Altitude MSE & Heading MSE \\
\midrule
Path A: direct MAE                       & 0.071 & 0.003 & 0.009 & 0.156 \\
Path B: voice $\rightarrow$ flow + head  & 0.710 & 0.012 & 0.022 & 1.998 \\
Path B: zero-shot (no head)              & 2.015 & 0.567 & 1.281 & 0.669 \\
\bottomrule
\end{tabular}
\end{table}

\begin{figure}[htbp]
\centering
\includegraphics[width=\textwidth]{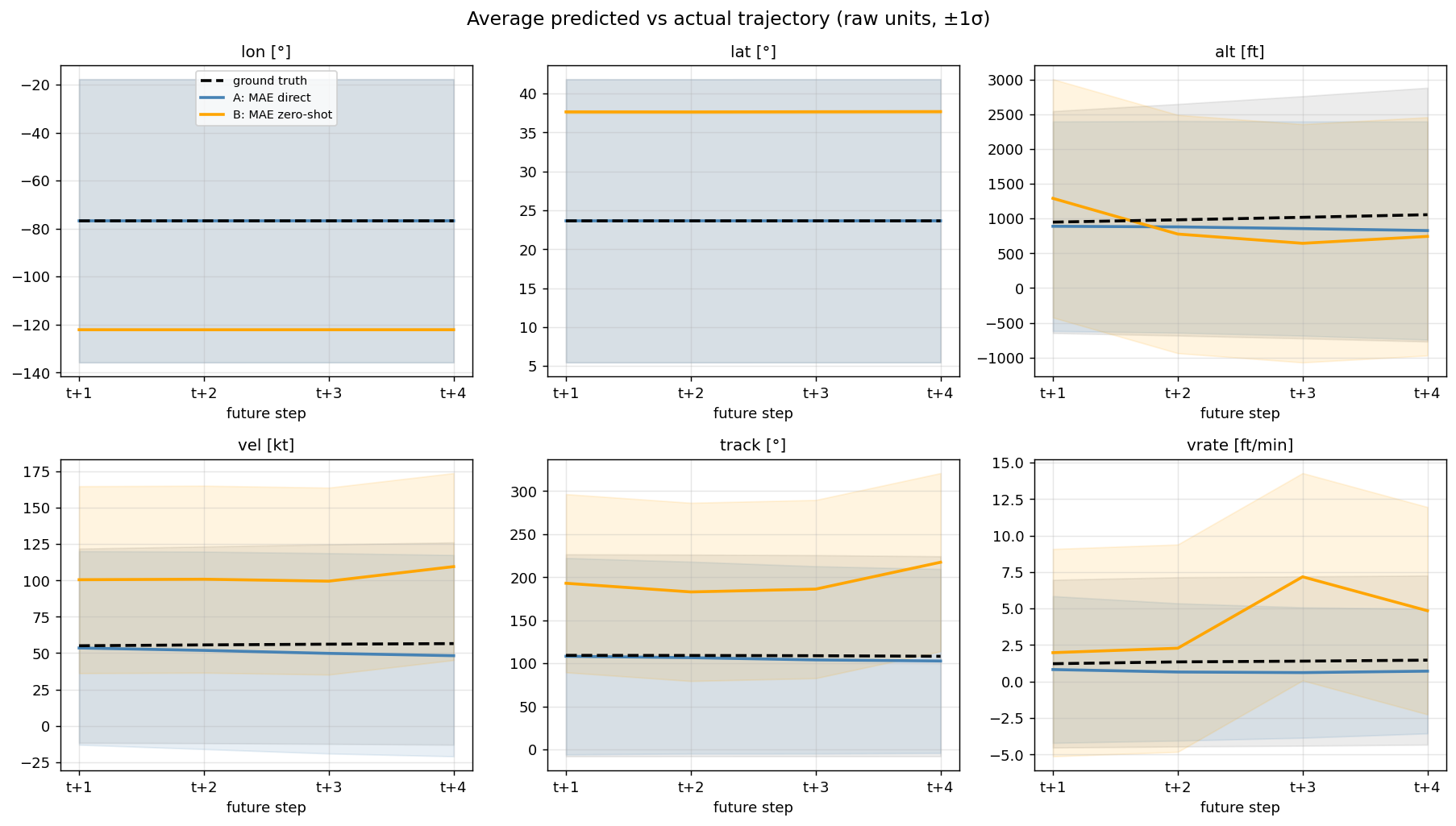}
\\[0.6em]
\includegraphics[width=\textwidth]{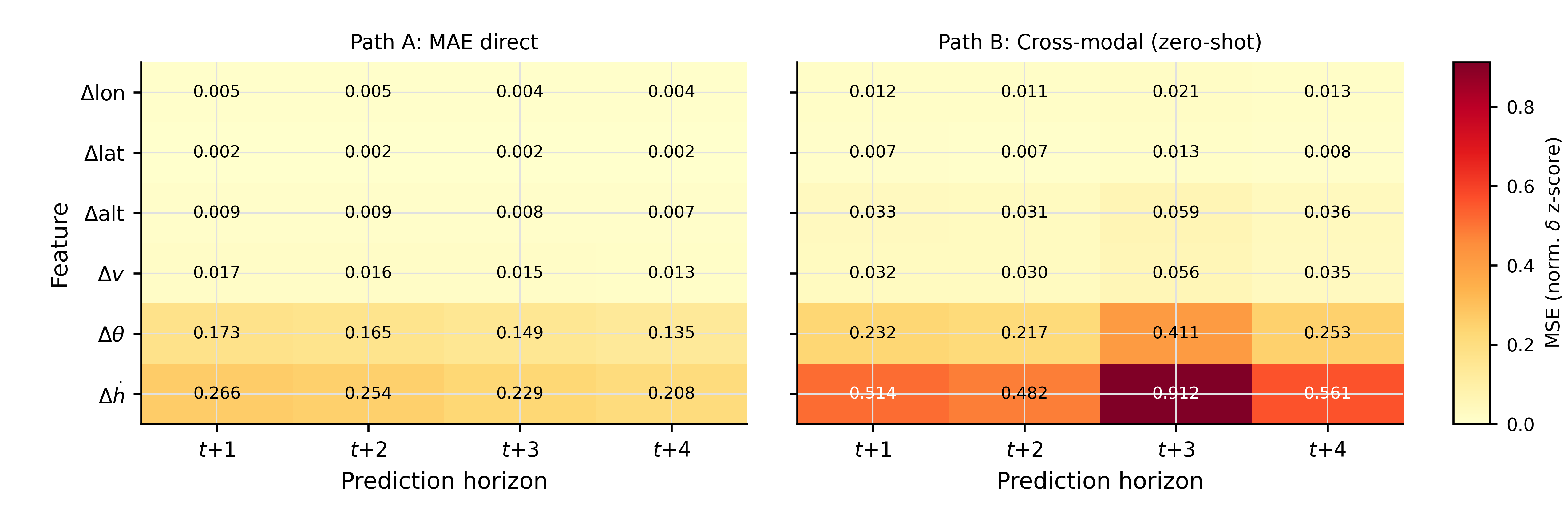}
\caption{Trajectory forecast results. \textit{Top}: average predicted profile against the ground truth over the full validation set, one sub-panel per feature (projected back into $\Delta$lon, $\Delta$lat, $\Delta$alt, $\Delta v$, $\Delta\theta$, $\Delta\dot{h}$). Path~A (direct MAE, solid) tracks the ground-truth mean closely, in line with its global MSE of $0.071$. Path~B (cross-modal chain, dashed) keeps the qualitative trend on position features but is offset on heading and vertical rate, and the zero-shot variant (dotted) deviates on all of them. \textit{Bottom}: per-feature and per-horizon MSE in normalized delta $z$-score space, for Path~A (\emph{left}) and for the zero-shot Path~B (\emph{right}). Heading ($\Delta\theta$) and vertical rate ($\Delta\dot{h}$) are the hardest targets for both paths. The error stays flat across the four horizons, which follows from the single-shot head.}
\label{fig:forecast}
\end{figure}

Results are reported in Table~\ref{tab:forecast} and illustrated in Fig.~\ref{fig:forecast}. Path~A reaches a global \emph{Mean Squared Error} (MSE) of $0.071$ in the normalized delta $z$-score space. Path~B reaches a global MSE of $0.710$, an order of magnitude higher. The gap is explained by the residual distribution shift between MAE latents and flow-reconstructed latents. The bijective stage matches paired points on average, but it does not equate the marginal distributions, so the downstream MAE reconstruction head, which was trained on real latents, is mildly out-of-distribution when fed flow-reconstructed ones. The naive zero-shot path that skips the StepPredictor degrades to an MSE of $2.0$, which confirms the importance of even a small task-specific head.

To make these numbers concrete, we convert the per-feature MSE to physical units using the Stage~1 per-feature normalization constants. The conversion is $\text{RMSE}_{\text{physical}} = \sqrt{\text{MSE}_z} \cdot \sigma_{\text{feature}}$ where, for the OpenSky San Francisco dataset, $\sigma_{\text{lon}} = 0.273^\circ$, $\sigma_{\text{lat}} = 0.292^\circ$, $\sigma_{\text{alt}} = \SI{1867}{\meter}$, and $\sigma_{\text{trk}} = 105.6^\circ$. At the Bay Area latitude $\phi \approx 37^\circ$\,N, one degree of longitude corresponds to about \SI{89}{\kilo\meter} and one degree of latitude to about \SI{111}{\kilo\meter}. For Path~A this gives an RMS position error of roughly \SI{1.5}{\kilo\meter} per predicted step, an RMS altitude error close to \SI{180}{\meter} ($\approx 580$\,ft), and an RMS heading error near $42^\circ$. For Path~B with the StepPredictor head the same conversion gives roughly \SI{3.1}{\kilo\meter}, \SI{280}{\meter} ($\approx 910$\,ft), and $86^\circ$. For the zero-shot variant the heading error reaches $149^\circ$ and altitude $\approx \SI{2.1}{\kilo\meter}$, which is essentially uninformative.

A per-feature breakdown shows that the first three quantities (longitude, latitude, altitude) are by far the easiest targets, while the next three (ground speed, true track, vertical rate) are the hardest. This split is natural, because positions are essentially integrated velocities, and since the model already observes the position trace, predicting the next delta is close to predicting the local velocity, which is itself close to an integration of what was just seen. The velocity-related quantities, by contrast, depend on acceleration, which is not directly observable from the input window. A controller's instruction such as ``climb and maintain flight level three five zero'' or ``turn left heading two seven zero'' can radically change either feature within a few seconds, and a surveillance-only forecast simply has no way to anticipate that change.

A second observation from Fig.~\ref{fig:forecast} is that the error does not grow with the horizon, and the per-step MSE stays roughly flat across the four predicted steps. This is a direct benefit of the single-shot prediction head, which outputs the four future steps in one forward pass instead of feeding its own output back as context. Autoregressive forecasters typically see an error budget that compounds with each step. The single-shot head trades this compounding for a slightly worse per-step accuracy at $t = 1$, but it keeps the four-step horizon stable. Longer horizons can still be obtained by repeating the prediction over successive windows, in which case the compounding behavior would re-appear and would have to be managed at the outer loop.

\subsection{Next Sentence Prediction}
\label{sec:results_nsp}
We also evaluate the reverse direction: given a trajectory window, retrieve the most likely voice phrase among the validation set. This is performed by ranking voice candidates by cosine similarity to $\Pt(\tremb)$ in the joint space. The retrieved phrase is then compared with the ground-truth transcription. The pattern of retrieval errors is informative. Aircraft on similar flight phases (e.g., climbing departures out of KSFO) retrieve voice phrases issued by the same controller and following the same phraseology, even when the exact addressee differs. This suggests that the joint space captures the coarse semantics of ATC instructions (instruction type, sector identity, flight phase) in addition to fine-grained identity matching. Per-tower retrieval breakdowns confirm that towers with stereotyped exchanges (ground, delivery) yield higher R@1 than towers with longer, more diverse instructions (NorCal departure), and that the cluster structure observed in the joint space (Section~\ref{sec:results_latent}) is consistent with this distinction.

\subsection{Ablation Study on the Joining Stage}
\label{sec:results_ablation}
\begin{figure}[bp]
\centering
\begin{subfigure}{0.48\textwidth}
  \includegraphics[width=\textwidth]{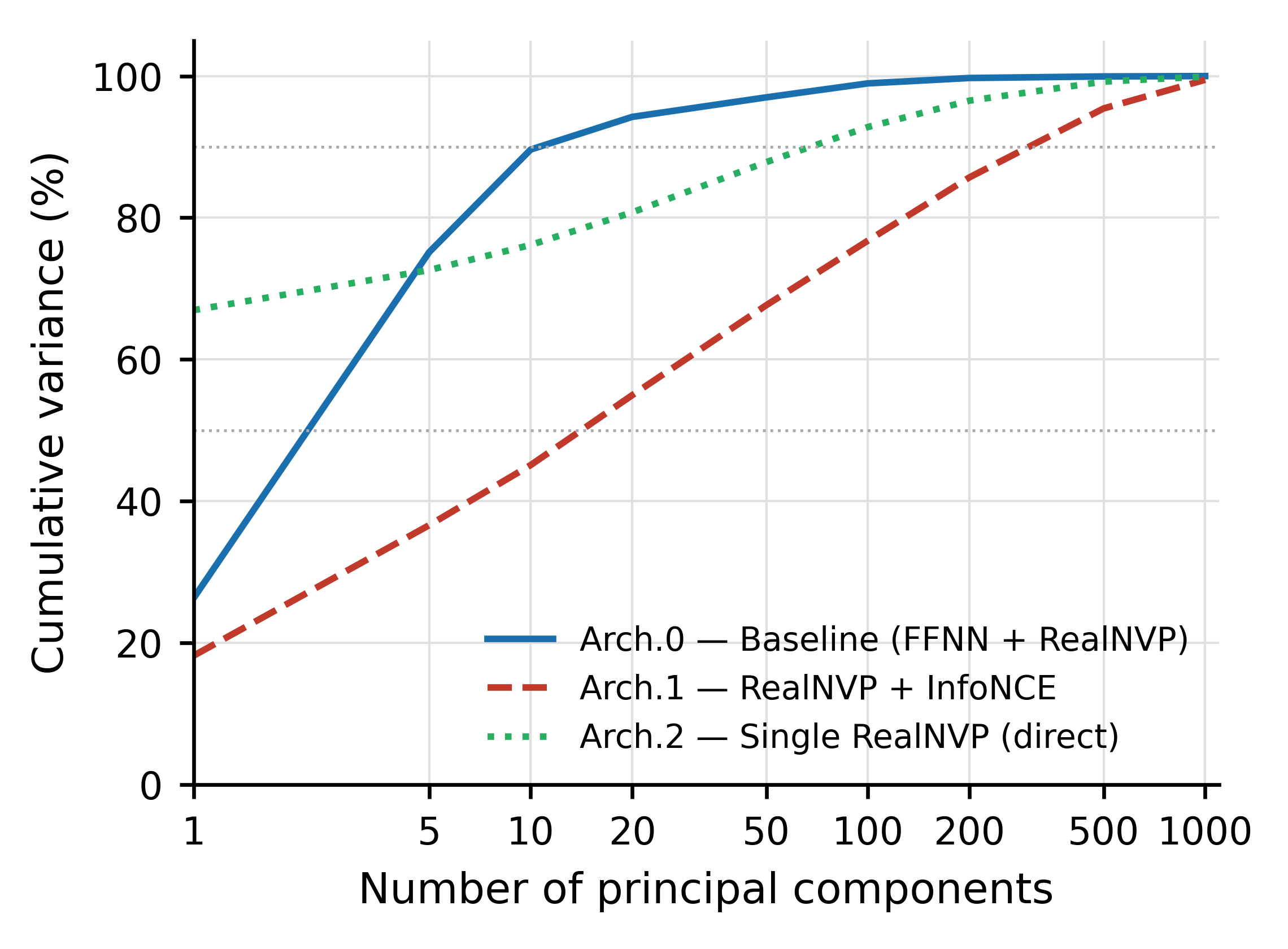}
  \caption{Cumulative variance carried by the sorted latent dimensions.}
\end{subfigure}\hfill
\begin{subfigure}{0.48\textwidth}
  \includegraphics[width=\textwidth]{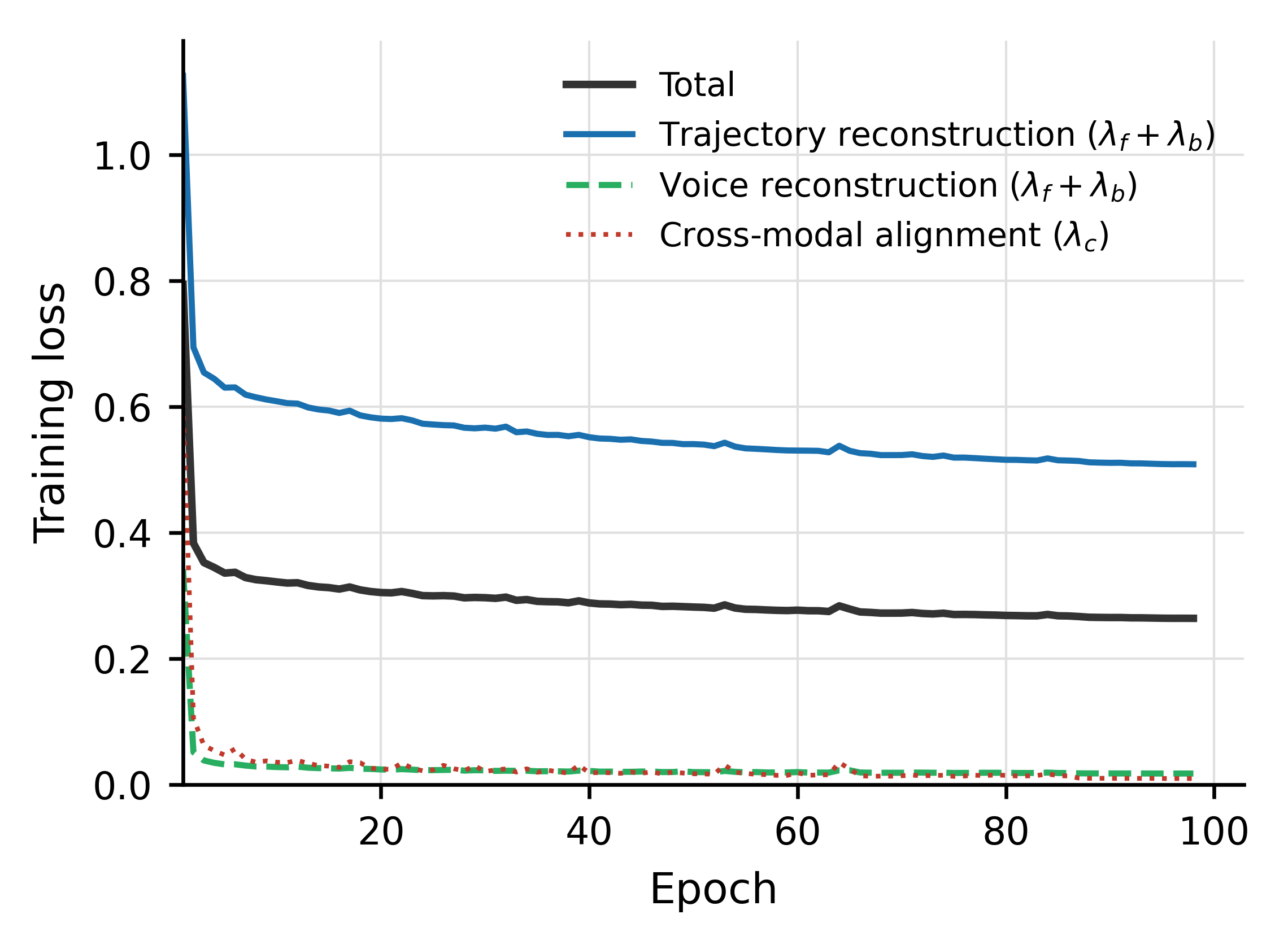}
  \caption{Per-term loss decomposition during stage~2 training.}
\end{subfigure}
\caption{Ablation of the joining stage, for the three architectures sketched in Fig.~\ref{fig:arch_variations}. (a)~Dotted horizontal lines mark the $50\%$ and $90\%$ variance thresholds. (b)~Baseline pipeline only, with the total loss split into its three terms.}
\label{fig:ablation}
\end{figure}

We compare the three alignment architectures sketched in Fig.~\ref{fig:arch_variations}. \emph{Arch.~0} is the baseline used throughout the article: a contrastive stage built on top of two FFNN projectors is followed by a bijective mimicking stage, that is, two normalizing flows trained with the loss of Eq.~\eqref{eq:nvp_loss}. \emph{Arch.~1} is a one-stage variant in which a single pair of normalizing flows directly plays both roles, and the same flows are trained jointly with the InfoNCE objective and a reconstruction term, without the FFNN projectors in between. \emph{Arch.~2} is a flow-only variant in which a single RealNVP is trained with cycle-consistency between the two modality embeddings, with no intermediate joint space and no contrastive supervision. It tests how much of the alignment can be recovered by reconstruction alone. The variance carried by each latent dimension of the joint space is reported in Table~\ref{tab:ablation} and the cumulative-variance profiles are plotted in Fig.~\ref{fig:ablation}.

\begin{table}[hbt!]
\caption{\label{tab:ablation}Ablation study on the joining stage. Cumulative variance carried by the first $K$ principal directions of the joint space, dimensionless, for the three architectures of Fig.~\ref{fig:arch_variations}.}
\centering
\begin{tabular}{lccc}
\toprule
Number of directions $K$ & Arch.~0 (baseline) & Arch.~1 (one-stage flow) & Arch.~2 (flow only) \\
\midrule
\hphantom{0}5  & 0.42 & 0.34 & 0.27 \\
10             & 0.61 & 0.51 & 0.43 \\
20             & 0.78 & 0.70 & 0.62 \\
50             & 0.93 & 0.88 & 0.84 \\
\bottomrule
\end{tabular}
\end{table}

Arch.~0 concentrates more variance in the first few principal directions of the joint space than the other two variants. Its first five directions carry about $42\%$ of the total variance against $27\%$ for the flow-only Arch.~2, and its first twenty carry $78\%$ against $62\%$. The effective rank of the joint space confirms the advantage of Arch.~0, at $203$ against $564$ for Arch.~2 and $806$ for Arch.~1. The two variants invert between the two metrics, since Arch.~1 concentrates slightly more variance in the leading directions than Arch.~2 but spreads the remainder over a markedly longer tail. This concentration is desirable, since a compact and coherent cross-modal alignment is one that uses a small number of axes consistently across both modalities. Contrastive supervision is what produces it, even when a separate flow is trained on top. Removing that supervision or folding it into the flow spreads the same information over many more directions. A second observation from the loss decomposition is that voice reconstruction converges roughly $35 \times$ faster than trajectory reconstruction. The asymmetry is structural. The voice embedding produced by Whisper concentrates most of its variance along a few principal directions of a semantically smooth manifold, while the flattened MAE embedding has high-dimensional correlations between timesteps that the flow must model simultaneously.

\subsection{Latent Space Analysis}
\label{sec:results_latent}
\begin{figure}[htbp]
\centering
\includegraphics[width=0.95\textwidth]{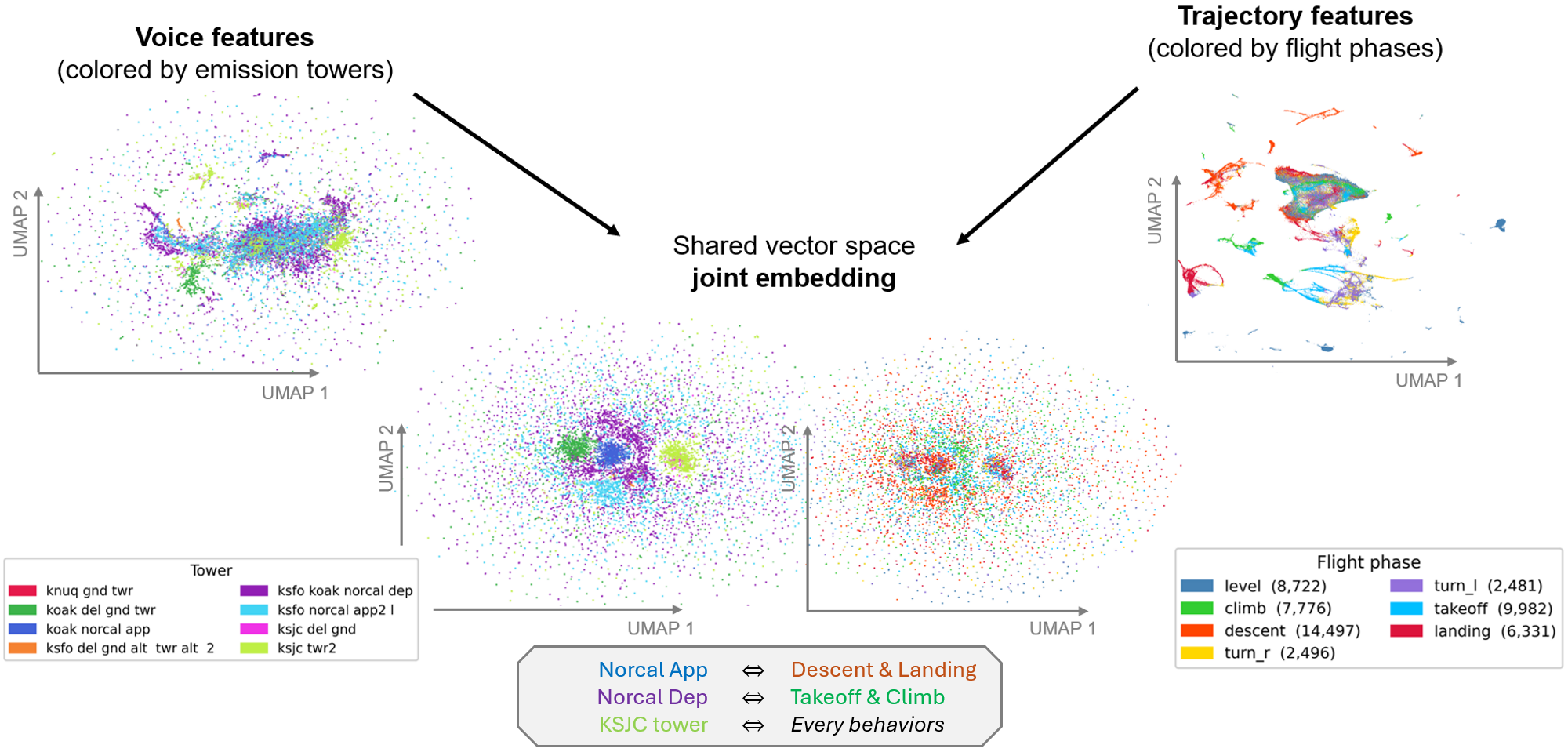}
\caption{UMAP projection of the joint space. The left panel shows the voice projections and the right panel the trajectory projections. The labels of the voice points are inferred from the voice signal itself, and the labels of the trajectory points are inferred from the kinematic features. The two modalities share the same coarse partition of the airspace, and flight-phase structure is visible inside each tower cluster.}
\label{fig:umap}
\end{figure}

To illustrate the structure of the learned space, we project voice and trajectory embeddings to two dimensions with \emph{Uniform Manifold Approximation and Projection} (UMAP)~\cite{mcinnes2018umap} and color the points by control tower and by flight phase. Figure~\ref{fig:umap} shows that the two modalities form overlapping clusters that organize primarily by tower (which corresponds to a coarse partition of the airspace) and secondarily by flight phase (ground vs.\ en-route). The fact that voice and trajectory projections share the same coarse structure, without ever being supervised on these labels, confirms that the contrastive objective recovers physically meaningful axes of variation.

A particularly striking observation is that ground frequencies (KOAK delivery, KSJC delivery) form tight, separated clusters from approach frequencies (NorCal approach, NorCal departure). This is consistent with what controllers report, namely that the phraseology and the typical aircraft state are very different on the two types of frequencies, and the joint space picks up the distinction without ever being told that towers exist as labels. Within each cluster, flight phase produces a secondary axis (ground taxi vs.\ climb vs.\ cruise) visible in the voice projection too, since the words used by controllers correlate with the kinematic regime of the aircraft they address. This double structure (tower + phase) is the empirical signature of a joint cartography of the airspace. We note that the joint space remains effectively low-dimensional even though we work in $\R^{1024}$, with the variance concentrated on a few principal directions (Section~\ref{sec:results_ablation}). This is the regime in which contrastive representation learning is known to be vulnerable to dimensional collapse~\cite{dimcollapse2021}, which is why the dropout and weight decay regularizers used during training, together with the bijective stage, are partly chosen to mitigate this risk.

\subsection{Continuity in the Trajectory Latent Space}
\label{sec:results_continuity}
\begin{figure}[tp]
\centering
\includegraphics[width=1\textwidth]{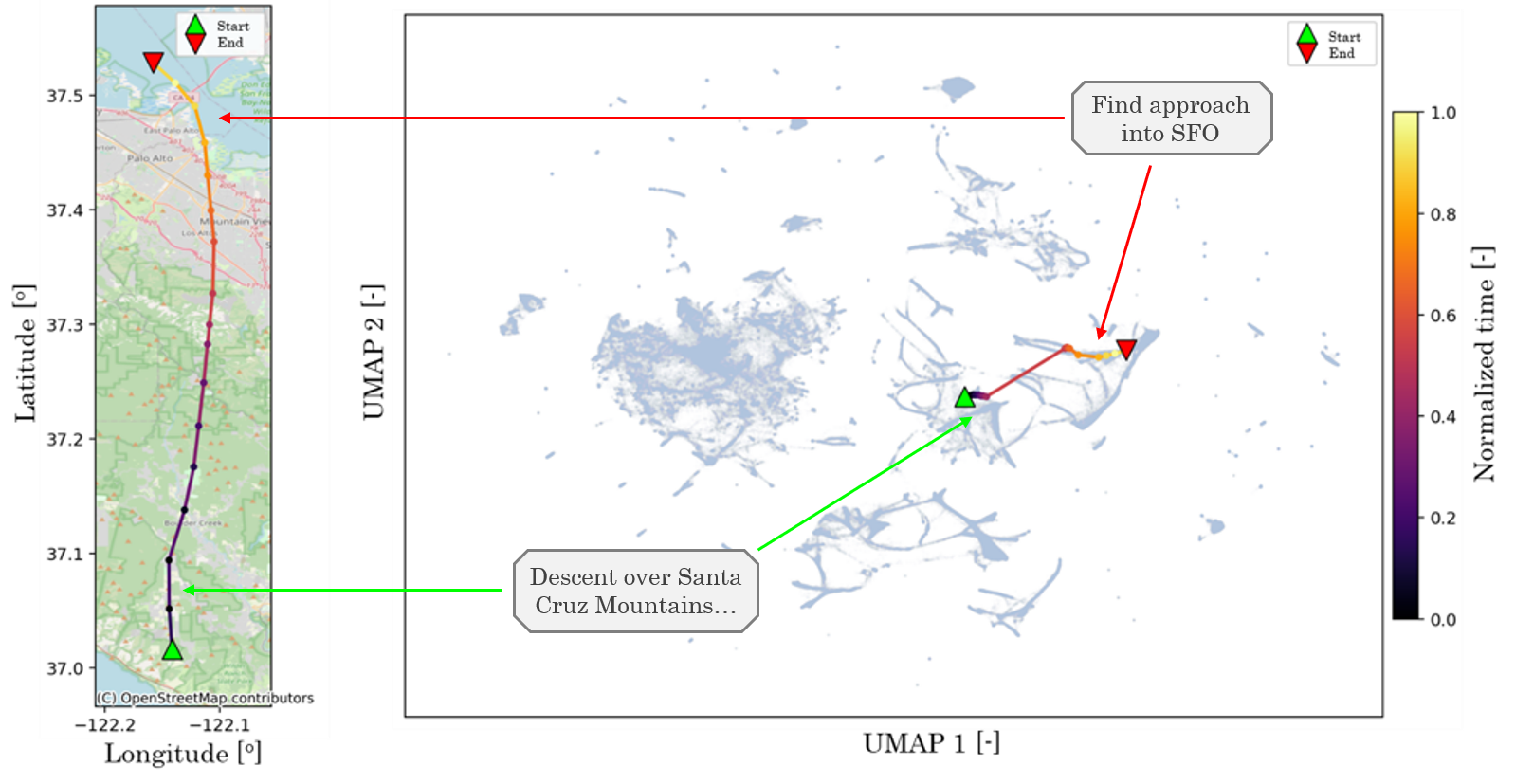}
\caption{Trajectory of a single aircraft in the latent space of the trajectory tower. The flight is continuous within each behavior cluster (descent, landing) and jumps when the flight phase changes. Clusters in this projection encode broad maneuver classes such as landing, climbing, or turning.}
\label{fig:continuity}
\end{figure}

Beyond the clustered structure analyzed in Section~\ref{sec:results_latent}, the trajectory latent space is also locally smooth. As an aircraft moves through successive trajectory windows, its representation traces a continuous path inside each cluster, and small changes in motion produce small displacements in the latent space. The trajectory only ``jumps'' from one cluster to another at flight-phase transitions, which can be read as a change of regime. Figure~\ref{fig:continuity} illustrates this behavior on a single inbound flight to KSFO. The aircraft starts in the descent cluster and moves continuously inside it, then jumps to the final approach and landing cluster as the flight phase changes from descent to touchdown, and moves continuously again until the trajectory ends. This piecewise-continuous structure mirrors the way a controller reasons about a flight, as a sequence of continuous regimes (climbing, cruising, descending, taxiing) separated by discrete maneuver transitions.

\section{Conclusion}
\label{sec:conclusion}
\subsection*{Achievements}

We presented a situational awareness framework that aligns ATC voice transmissions and aircraft trajectories in a single learned latent space. V2TATC follows a two-tower design with two complementary training stages: a contrastive alignment of pre-trained per-modality embeddings, followed by a bijective lifting through normalizing flows. We have shown that this framework can be instantiated over a congested terminal area using public data only, that the resulting joint space supports cross-modal retrieval well above random, and that it carries enough structure to drive simple downstream tasks such as short-horizon trajectory forecasting from voice instructions.

Several practical uses can be envisioned. The most immediate one is a controller-facing visualization in which a click on an aircraft retrieves the recent voice exchanges that concern it, and conversely. A second use is anomaly detection, where voice instructions and observed trajectories that are too far apart in the joint space could signal a non-compliant maneuver or a developing situational-awareness mismatch. A third direction, which we leave to future work, is to leverage the bijective mapping to generate synthetic voice--trajectory pairs for stress-testing other decision support tools, or to drive a more capable language model conditioned on the surveillance picture. V2TATC's framework is generic. It does not depend on the specific encoders we used, and could be extended to additional modalities such as flight-plan data.

The current limitations are also visible. Path~B (voice-conditioned trajectory forecast) is still an order of magnitude less precise than Path~A, mainly because of the residual distribution shift introduced by the bijective stage. Three directions could close this gap. Adversarial alignment between MAE latents and flow-reconstructed latents would attack the distribution shift directly. Fine-tuning the StepPredictor end-to-end with the flow would instead let the head adapt to the latents it actually receives. Collecting data over additional terminal areas would scale the contrastive dataset and reduce the residual overfitting observed here. We expect that the combination of these three directions can bring Path~B within a factor of two of Path~A, which would make the framework operationally relevant.

\subsection*{Future Work}

We plan to keep developing the framework along three complementary axes. First, we will focus on the voice-to-trajectory direction. We want to design more expressive joining architectures that better capture the variability of the airspace. In particular, we plan to apply classifier-free guidance and diffusion models, so that voice instructions can act as guidance signals for a generative trajectory model. Second, we will broaden the set of input modalities that the framework can ingest. The current paired dataset is already synchronized with the METAR observations collected during the same four-day window, so weather data can be added without additional alignment work. Flight-plan and METAR streams could then be combined with voice and trajectories in a single multimodal joint space. Third, we will continue scraping new data and improve the post-processing pipeline. The goal is to grow the paired dataset to a size where the contrastive data samples the embedding space more uniformly, and where the joint dataset reflects the diversity of operations across multiple terminal areas rather than the San Francisco Bay Area alone.

\section*{Funding Sources}
This work was supported by Intelmatix and Prince Sultan University (PSU). Louis Brusset was hosted at CITRIS and the Banatao Institute, University of California, Berkeley, under the Visiting Scholar Researcher program.

\section*{Acknowledgments}
The authors would like to thank Professor Trevor Darrell from the Berkeley Artificial Intelligence Research (BAIR) Lab for his early conversations on joint embeddings. We also would like to thank Tom Davis, Jim Murphy and Vishwanath Bulusu from Crown Innovations LLC, John Robinson and Parimal Kopardekar of NASA Ames Research Center, and Dragos Margineantu from Boeing for their insightful discussions throughout the project. We are grateful to LiveATC.net for granting access to the ATC audio archives that made this study possible. We also thank the OpenSky Network and aviationweather.gov for openly sharing their surveillance and weather data, and more broadly for their sustained data-collection initiatives, which are essential to research such as ours.

\bibliography{references}

\section*{Appendix}
The appendix collects supporting material that complements the main article and is useful for reproduction.

\subsection*{A. Hyperparameters}
The contrastive projectors are single-hidden-layer FFNNs of width $4096$ with SELU activation~\cite{klambauer2017selu} and a dropout of $0.2$. The Scaled Exponential Linear Unit was selected over more common activations such as ReLU because it is smooth, differentiable everywhere, and, when combined with the associated LeCun-normal initialization, preserves the mean and variance of its input across layers, so activations remain approximately zero-mean and unit-variance throughout the projector. This self-normalization property reduces the need for explicit batch normalization inside the projector and stabilizes contrastive training when the batch size is limited. The InfoNCE temperature is initialized at $\tau = 0.07$, parameterized in log-space, and clamped to $[0.04, 100]$. Training uses AdamW ($\beta_1 = 0.9$, $\beta_2 = 0.999$), gradient clipping at norm $1.0$, batch size 512, learning rate $10^{-4}$ with \texttt{ReduceLROnPlateau} scheduling, and early stopping on the validation loss with a patience of fifteen epochs.

The two normalizing flows are stacks of eight affine coupling layers with conditioner MLPs of hidden width $[256, 256]$. ActNorm layers~\cite{kingma2018glow} are inserted between coupling blocks and initialized data-dependently. Smaller-dimension embeddings are zero-padded to the common working width $D_\text{max} = 1792$ before entering the flow. The bijective stage uses the same optimizer with batch size 256 and a maximum of 200 epochs. The MAE encoder is a four-layer Transformer with hidden width $d = 128$, four attention heads, and feed-forward dimension $512$. Training uses AdamW, peak learning rate $10^{-3}$, linear warm-up over five epochs followed by cosine decay, weight decay $10^{-2}$ (excluding bias and normalization layers), gradient clipping at $1.0$, batch size 1024, and mixed-precision training. The masking ratio is $m = 0.25$ for Stage~1 and configurable for Stage~2.

\subsection*{B. Trajectory Encoder Backbones}
We compared four backbone architectures inside the MAE wrapper described in Section~\ref{sec:approach}: a Transformer~\cite{vaswani2017attention}, a bidirectional LSTM, a scalar xLSTM (sLSTM cells only), and a full xLSTM with matrix memory~\cite{beck2024xlstm}. Validation losses on the OpenSky San Francisco dataset are reported in Table~\ref{tab:mae_results}. All four backbones converge to validation MSE values within $4 \times 10^{-4}$ of one another, which indicates that the MAE objective on 14-timestep ADS-B windows is largely architecture-agnostic at this scale.

\begin{table}[hbt!]
\caption{\label{tab:mae_results}Validation MSE (dimensionless, normalized-delta space) for four trajectory backbones trained as MAE on Bay Area ADS-B data. All values are within $4 \times 10^{-4}$ of one another; the Transformer is retained as the default encoder.}
\centering
\begin{tabular}{lcc}
\toprule
Backbone & Validation MSE & Relative training cost \\
\midrule
Transformer (4 layers, $d=128$) & 0.0351 & $1.0\times$ \\
Bidirectional LSTM              & 0.0353 & $0.23\times$ \\
Scalar xLSTM (sLSTM only)       & 0.0354 & $1.2\times$ \\
Full xLSTM (sLSTM + mLSTM)      & 0.0354 & $3.7\times$ \\
\bottomrule
\end{tabular}
\end{table}

\begin{figure}[htbp]
\centering
\includegraphics[width=0.95\textwidth]{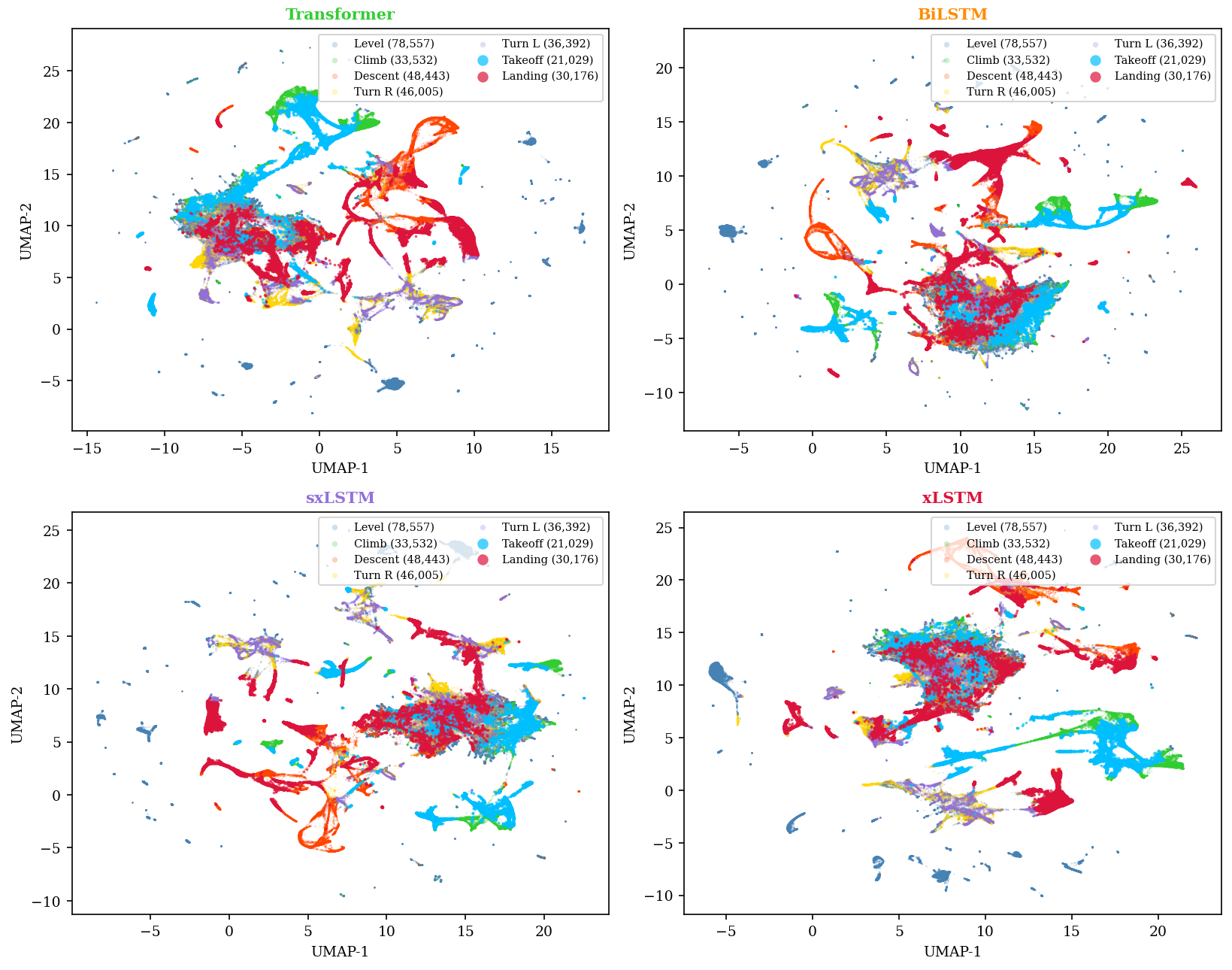}
\caption{UMAP projections of the MAE trajectory latent space for the four candidate backbones (Transformer, BiLSTM, sLSTM, xLSTM). Colors indicate flight phase labels inferred directly from the raw kinematic features.}
\label{fig:umap_phases}
\end{figure}

The UMAP projections of the encoder hidden states (Fig.~\ref{fig:umap_phases}) confirm this quantitative similarity. Beyond the near-identical validation MSE reported in Table~\ref{tab:mae_results}, the four latent spaces exhibit the same qualitative structure: similar pairs of neighboring colors, a large central blob, and the same filament-like extensions toward the periphery. This is a strong signal on its own, as it indicates that the geometry of the latent space captures an intrinsic property of the airspace itself and does not depend on the specific inductive bias of the sequence model. Given this near equivalence, we retain the Transformer because (i) it has the lowest validation MSE of the four candidates and (ii) its main blob is qualitatively the most spread out and the most cleanly organized by flight phase. We note that BiLSTM is about four times cheaper to train per epoch (Table~\ref{tab:mae_results}), which would be a relevant trade-off in a larger-scale experiment. The standalone UMAP of the Transformer latent space is shown in the body of the article as Fig.~\ref{fig:transformer_proj}.

The projections carry a few observations that are physically interpretable and that are consistent across all four backbones. The large central blob is populated almost exclusively by general aviation trajectories. It is continuous, connects all flight phases (taxi, climb, cruise, descent, landing) into a single manifold, and reflects the fact that general aviation traffic follows fewer standardized procedures than commercial traffic. Commercial-aviation trajectories, by contrast, form tighter and more clearly separated clusters, because commercial flights follow standard instrument procedures and standard arrival and departure routes that repeat from one flight to the next. Local continuity between neighboring clusters is also visible and follows the temporal ordering of a flight: landing clusters sit next to descent clusters, and take-off clusters sit next to climb clusters, so a real flight traces a nearly continuous path across the latent space as it transitions from one phase to the next. Several disjoint clusters can be observed for a single flight phase; these correspond to the geographic separation between the main Bay Area airports (KSFO, KOAK, KSJC, KNUQ), each of which produces its own family of approach and departure trajectories despite sharing the same phase label.

\subsection*{C. Choice of the Voice Encoder}
\begin{figure}[htbp]
\centering
\begin{subfigure}{0.48\textwidth}
  \includegraphics[width=\textwidth]{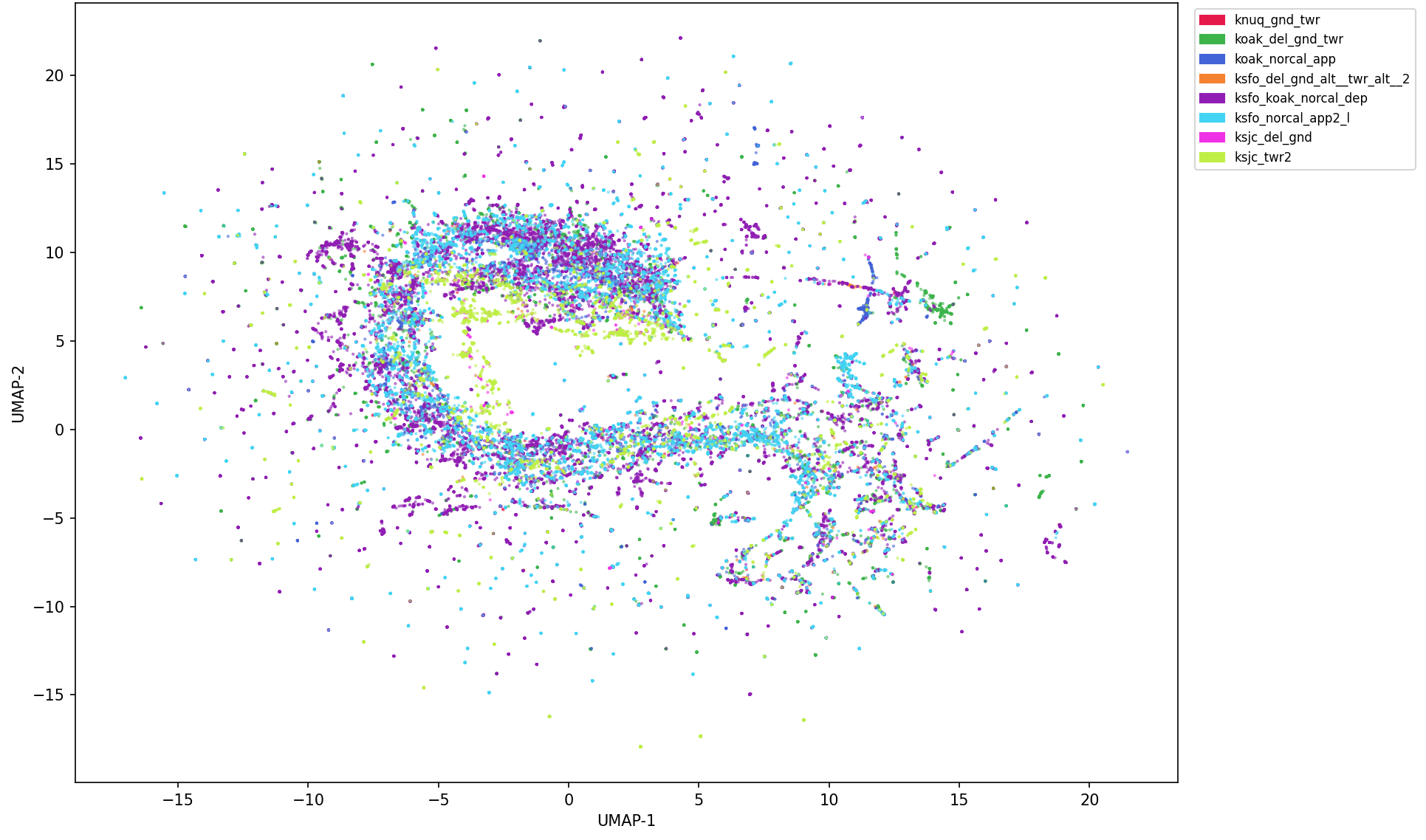}
  \caption{Whisper large-v3 embeddings.}
\end{subfigure}\hfill
\begin{subfigure}{0.48\textwidth}
  \includegraphics[width=\textwidth]{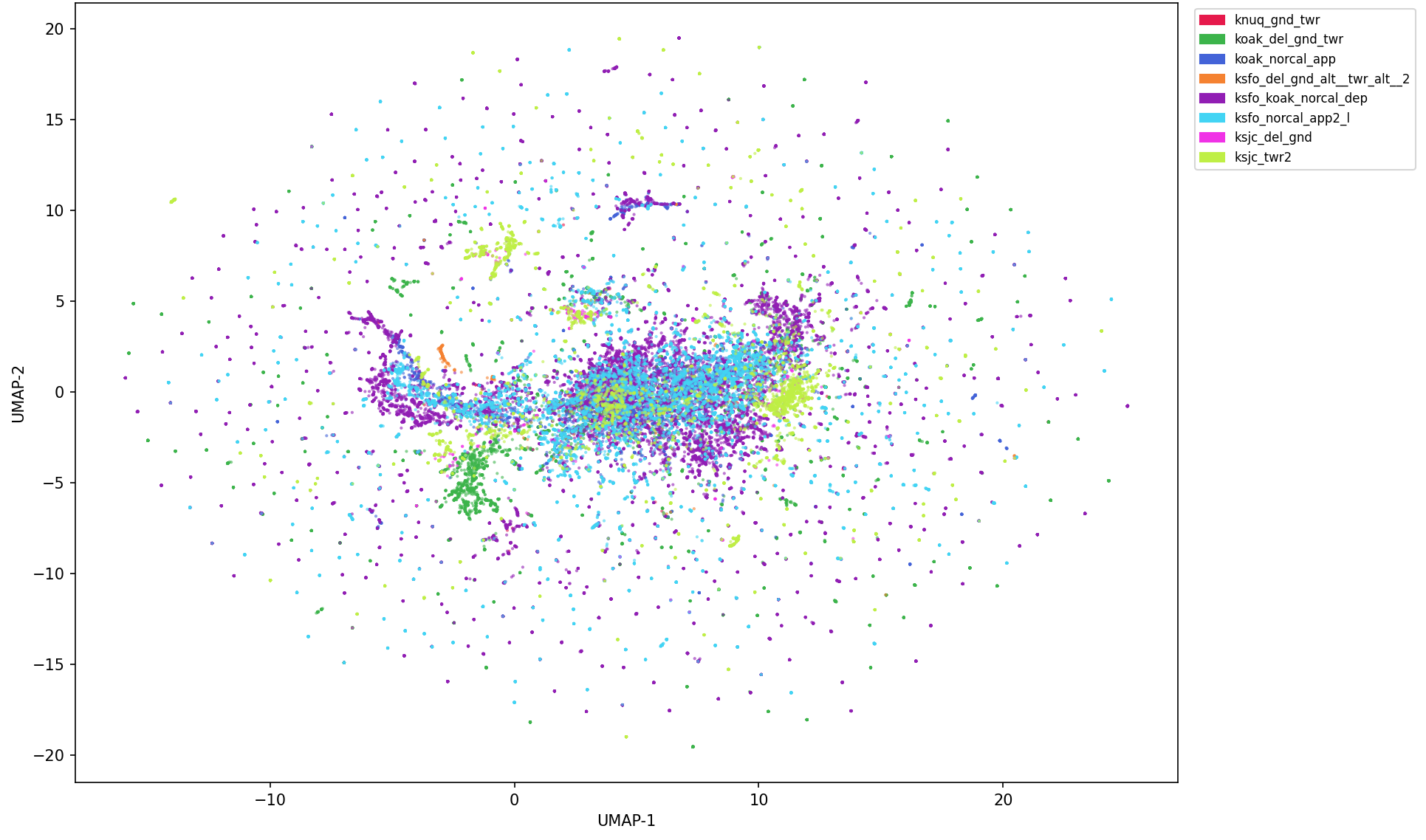}
  \caption{Wav2Vec~2.0 XLS-R embeddings (ATC fine-tuned).}
\end{subfigure}
\caption{UMAP projections of the voice embeddings, colored by control tower. Whisper produces a smoother and more spread-out latent space, while Wav2Vec~2.0 produces tighter but less smooth clusters.}
\label{fig:voice_umaps}
\end{figure}

We considered two foundation ASR models as voice encoders: Whisper large-v3~\cite{radford2023whisper} and Wav2Vec~2.0 XLS-R fine-tuned on an ATC corpus~\cite{baevski2020wav2vec2,zuluaga2022atco2}. The two models produce qualitatively different latent spaces. Figure~\ref{fig:voice_umaps} shows the UMAP projections of the voice embeddings, colored by control tower. Three visual properties of these projections drive our choice in favor of Whisper.

First, the Whisper latent space is markedly more \emph{spread out}. It occupies a wider portion of the UMAP plane and takes an elongated, filament-like shape that spans several visibly connected regions. The Wav2Vec~2.0 latent space, by contrast, collapses into a more compact blob with a few isolated satellites. A wider effective support is desirable for the downstream contrastive stage because the projector $\Pv$ has more room to separate distinct utterances without having to fight against a pre-existing collapse.

Second, the Whisper latent space is more \emph{locally smooth}, so that neighboring points in UMAP correspond to acoustically neighboring utterances, and clusters connect through continuous transition regions rather than through empty gaps. Smoothness matters directly for contrastive training, because a small perturbation of the input produces a small perturbation of the embedding, which yields well-behaved gradients through $\Pv$ and stable in-batch negatives. The Wav2Vec~2.0 projection, although it groups points by tower more cleanly on a coarse scale, is granular inside each cluster, where tight clusters are made of many small islands rather than a smooth density, so nearby points can correspond to unrelated utterances and the effective Lipschitz constant of the encoder is worse.

Third, the tighter clustering of Wav2Vec~2.0 is a direct consequence of the ATC fine-tuning. The model has specialized so strongly on the ATC domain that it has partially lost the broad acoustic prior it was pre-trained with, a phenomenon known as catastrophic forgetting. Because our joint dataset is small compared with the corpora on which foundation ASR models are trained, this loss is not compensated by our own data. Whisper, kept frozen, retains the full acoustic prior of its \SI{680000}{\hour} of multi-domain training, which turns out to be more valuable than the domain-specific but narrower Wav2Vec~2.0 representation.

Taken together, the three properties suggest that the Whisper representation is the better substrate for a joint embedding: a wider support, smoother local geometry, and a stronger acoustic prior. This visual analysis is consistent with the two other criteria discussed below (transcription quality when decoding back from the embedding, and validation loss under the encoder--projector sweep of Table~\ref{tab:sweep}), which independently favor Whisper.

A second decisive observation concerns the quality of transcriptions decoded back from the embedding. When the Whisper embedding is passed through a Whisper decoder, the recovered text is fluent and grammatically correct. When the Wav2Vec~2.0 embedding is decoded back to text, the output is not even a grammatically valid English sentence. Since our downstream goal includes recovering plausible ATC phraseology from a point in the joint space, this observation alone is enough to retain Whisper.

\begin{figure}[bp]
\centering
\includegraphics[width=0.75\textwidth]{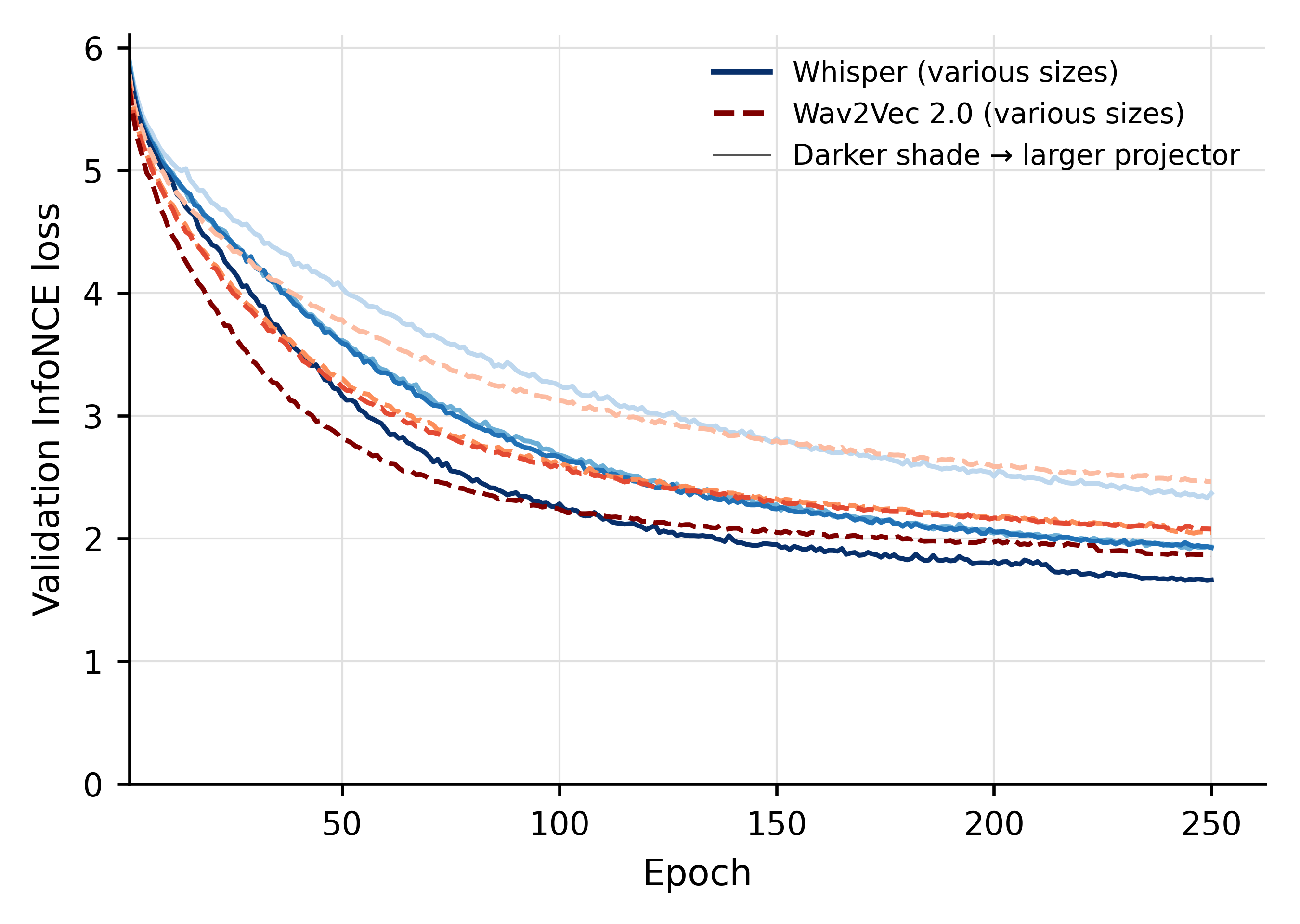}
\caption{Validation InfoNCE loss during contrastive training for the eight configurations of the encoder--projector sweep (two voice encoders $\times$ four projector sizes). Whisper large-v3 (solid lines, blue shades) consistently achieves a lower validation loss than Wav2Vec~2.0 (dashed lines, red shades) across all projector configurations. Performance saturates once the projector output dimension reaches $d_j = 1024$, which motivates our working configuration. Darker shades correspond to larger projector hidden dimensions.}
\label{fig:sweep}
\end{figure}

The two observations together support a general conclusion. In principle, a model fine-tuned on the target domain should outperform a generic foundation model. In practice, ATC fine-tuning is performed on relatively small corpora, and catastrophic forgetting of the broad acoustic knowledge embedded in the original model is hard to avoid. Foundation models like Whisper, trained on huge and acoustically diverse corpora (including noisy radio-style speech and many accents), turn out to be more robust on our data. To quantify this choice and to pick the projector width, we ran a small grid over voice encoders and projector dimensions. Results are summarized in Table~\ref{tab:sweep} and Fig.~\ref{fig:sweep}. Whisper outperforms Wav2Vec~2.0 at every projector size, and the validation loss is essentially flat once the projector output dimension reaches $1024$. We therefore use a single-hidden-layer projector with $d_j = 1024$ as the working configuration, which keeps the projector at the smallest width that saturates the validation loss, as discussed in Section~\ref{sec:implementation}.

\begin{table}[hbt!]
\caption{\label{tab:sweep}Validation metrics for the voice-encoder/projector-size sweep. Whisper consistently outperforms Wav2Vec~2.0 across all projector sizes, and the largest projector $(4096 \to 2048)$ gives the best retrieval.}
\centering
\begin{tabular}{lcccc}
\toprule
Configuration & Val loss $\downarrow$ & R@1 $\uparrow$ & R@5 $\uparrow$ & R@10 $\uparrow$ \\
\midrule
Whisper $(4096 \to 2048)$       & \textbf{1.5975} & \textbf{0.2423} & \textbf{0.6130} & \textbf{0.7364} \\
Whisper $(2048 \to 2048)$       & 1.6210 & 0.2387 & 0.5977 & 0.7263 \\
Whisper $(2048 \to 1024)$       & 1.6135 & 0.2391 & 0.6000 & 0.7297 \\
Whisper $(1024 \to 1024)$       & 1.7465 & 0.2177 & 0.5643 & 0.6938 \\
Wav2Vec~2.0 $(4096 \to 2048)$   & 1.8295 & 0.2163 & 0.5545 & 0.6758 \\
Wav2Vec~2.0 $(2048 \to 2048)$   & 1.8553 & 0.2119 & 0.5534 & 0.6781 \\
Wav2Vec~2.0 $(2048 \to 1024)$   & 1.8465 & 0.2114 & 0.5511 & 0.6731 \\
Wav2Vec~2.0 $(1024 \to 1024)$   & 1.9912 & 0.1909 & 0.5117 & 0.6423 \\
\bottomrule
\end{tabular}
\end{table}

\subsection*{D. Affine Coupling Layer in RealNVP}
\begin{figure}[htbp]
\centering
\includegraphics[width=1\textwidth]{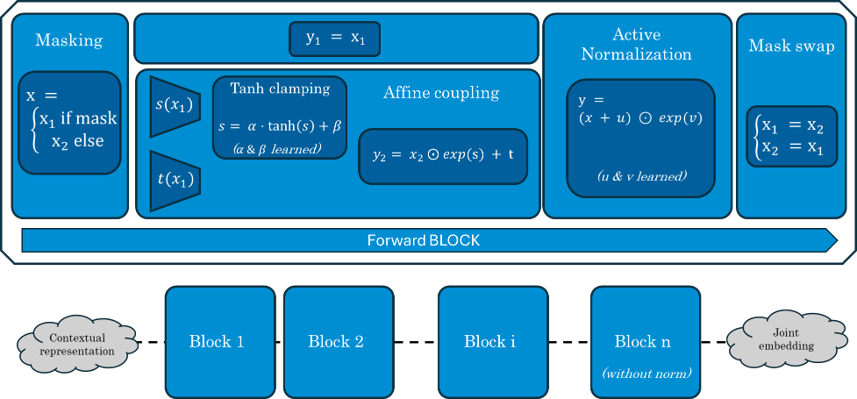}
\caption{Schematic of a RealNVP affine coupling layer. The input is split in two halves. One half is left unchanged and used as the conditioning input of two neural networks that produce the scale $s$ and translation $t$ of an affine transformation applied to the other half. The inverse can be computed analytically in closed form. These affine coupling layers are then stacked, with the mask swapped by half at each block, so that the flow spreads across all dimensions.}
\label{fig:realnvp_explained}
\end{figure}

For completeness, Fig.~\ref{fig:realnvp_explained} gives a self-contained schematic of the forward and inverse passes of an affine coupling layer in a RealNVP flow. The same idea is repeated $K$ times with alternating masks to build the two normalizing flows $\fv$ and $\ft$ used in the bijective stage of the framework.

\subsection*{E. Code Repositories and Dataset}
The dataset is released as sharded HDF5 files organized per control tower. Each shard exposes four parallel datasets: variable-length \texttt{int16} \texttt{waveforms} at \SI{16}{\kilo\hertz}; \texttt{float32} \texttt{trajectories} of shape $(N, T, 7)$ where the first six columns are longitude, latitude, geometric altitude, ground speed, true track, and vertical rate, and the seventh carries the absolute timestamp; per-sample \texttt{metadata\_json} strings containing callsign, phrase timestamps, ICAO~24 code, match type, and source MP3 filename; and global \texttt{sample\_indices}. The code for collection, preprocessing, analysis, and training is split across five public repositories:

\begin{itemize}
  \item \href{https://github.com/LouisBrusset/CITRIS-llm-for-atc-dataset}{\texttt{LouisBrusset/CITRIS-llm-for-atc-dataset}} -- audio-archive ingestion from LiveATC.net and ADS-B polling from OpenSky.
  \item \href{https://github.com/LouisBrusset/CITRIS-joint-embedding-dataset-preprocessing}{\texttt{LouisBrusset/CITRIS-joint-embedding-dataset-preprocessing}} -- transcription, callsign matching, HDF5 assembly.
  \item \href{https://github.com/LouisBrusset/CITRIS-joint-dataset-analysis}{\texttt{LouisBrusset/CITRIS-joint-dataset-analysis}} -- statistical analysis and wind-field estimation (additional study).
  \item \href{https://github.com/matu1003/CITRIS-modern-transformer-encoding}{\texttt{matu1003/CITRIS-modern-transformer-encoding}} -- MAE pre-training of the trajectory encoder.
  \item \href{https://github.com/LouisBrusset/CITRIS-joint-embedding-Voice2Traj}{\texttt{LouisBrusset/CITRIS-joint-embedding-Voice2Traj}} -- contrastive projectors, RealNVP bijective flows, and downstream evaluation.
\end{itemize}

A subset of the dataset is also released on HuggingFace for training and experimentation at the following link: \href{https://huggingface.co/datasets/Lbrusset/SF_bay_voice2traj_dataset}{\texttt{Lbrusset/SF\_bay\_voice2traj\_dataset}}. The subset contains approximately 83{,}223 paired audio--trajectory samples together with the metadata describing how each pair was linked (callsign match type, timestamps, ICAO~24, source frequency, and source MP3 file).

\end{document}